\documentclass[10pt,twocolumn,letterpaper]{article}

\usepackage{iccv}
\usepackage{diagbox}
\usepackage{times}
\usepackage{epsfig}
\usepackage{graphicx}
\usepackage{multirow}
\usepackage{amsmath}
\usepackage{amssymb}
\usepackage{color}
\usepackage{booktabs} 
\usepackage{nicefrac}
\usepackage{footnote}
\usepackage{multicol}

\usepackage[percent]{overpic}
\usepackage{xcolor,colortbl}
\usepackage{float}
\restylefloat{table}

\definecolor{gold}{rgb}{0.81,0.71,0.23} 
\definecolor{silver}{rgb}{0.75,0.75,0.75}
\definecolor{bronze}{rgb}{0.69,0.55,0.34}
\newcommand{\rankfirst}[1]{{\cellcolor{gold} #1}}
\newcommand{\ranksecond}[1]{{\cellcolor{silver} #1}}
\newcommand{\rankthird}[1]{{\cellcolor{bronze} #1}}

\usepackage[pagebackref=true,breaklinks=true,colorlinks,bookmarks=false]{hyperref}

\usepackage{cleveref}

\iccvfinalcopy 

\def\iccvPaperID{7821}

\newcommand\tableentry[2]{#1, #2\textdegree} 
\newcommand\tableentrytra[1]{#1, ~-~~~~}

\def\etal{{et~al.}}

\def\eg{{\it e.g.,\ }}
\def\vs{{\it vs.\ }}
\def\cf{{\it cf.,\ }}

\definecolor{turquoise}{cmyk}{0.65,0,0.1,0.1}
\definecolor{purple}{rgb}{0.65,0,0.65}
\definecolor{dark_green}{rgb}{0, 0.5, 0}
\definecolor{orange}{rgb}{0.8, 0.2, 0.2}
\definecolor{red}{rgb}{1, 0, 0}
\definecolor{inpaper}{rgb}{0.5, 0.5, 0.5}

\iftrue
\newcommand{\ozgur}[1]{\textcolor[rgb]{0.1,0.5,0.5}{\textbf{O: #1}}}  
\newcommand{\aron}[1]{\textcolor[rgb]{0.35,0.56,0.41}{\textbf{\small #1 [A]}}}
\newcommand{\eric}[1]{\textcolor[rgb]{0.5,0.2,0.5}{\textbf{E: #1}}}
\newcommand{\gb}[1]{\textcolor[rgb]{0.1,0.8,0.2}{\textbf{G: #1}}}
\newcommand{\konrad}[1]{\textcolor[rgb]{0.1,0.2,0.5}{\textbf{K: #1}}}
\newcommand{\todo}[1]{{\color{red} \bf [TODO: #1]}}
\newcommand{\old}[1]{{\color{red} [OLD: #1]}}
\newcommand{\iccv}[1]{{\color{purple} #1}}
\else
\newcommand{\ozgur}[1]{}  
\newcommand{\aron}[1]{} 
\newcommand{\eric}[1]{}
\newcommand{\gb}[1]{}
\newcommand{\konrad}[1]{}
\newcommand{\todo}[1]{}
\newcommand{\old}[1]{}
\newcommand{\iccv}[1]{#1}
\fi

\newcommand{\STAB}[1]{\begin{tabular}{@{}c@{}}#1\end{tabular}}

\usepackage{textcomp}

\newcommand{\textapprox}{\raisebox{0.5ex}{\texttildelow}}

\usepackage{refcount}

\newcommand{\myfootnotetext}[1]{\footnotetext{#1\label{fn:text}
        \edef\fnmark{\getpagerefnumber{fn:mark}}
        \edef\fntext{\getpagerefnumber{fn:text}}
        \ifx\fnmark\fntext\else\ClassWarning{}{footnote mark and text on different pages!}\fi}}

\newif\ifmyarxiv
\myarxivtrue

\ificcvfinal\fi
\begin{document}

\title{Visual Camera Re-Localization \\ Using Graph Neural Networks and Relative Pose Supervision}

\author{Mehmet Ozgur Turkoglu\textsuperscript{1}\thanks{Work done during an internship at Niantic.}\ \ 
\and
Eric Brachmann\textsuperscript{2}\ \ \ 
\and
Konrad Schindler\textsuperscript{1}\ \ \ 
\and
Gabriel J. Brostow\textsuperscript{2,3}\ \ \  \\
\and
Aron Monszpart\textsuperscript{2} \ \ \  \\
\and
\textsuperscript{1}ETH Zurich \hspace{1cm} \textsuperscript{2}Niantic \hspace{1cm} \textsuperscript{3}University College London
}

\maketitle

\begin{abstract}
Visual re-localization means using a single image as input to estimate the camera's location and orientation relative to a pre-recorded environment. The highest-scoring methods are ``structure-based,'' and need the query camera's intrinsics as an input to the model, with careful geometric optimization. When intrinsics are absent, methods vie for accuracy by making various other assumptions. This yields fairly good localization scores, but the models are ``narrow'' in some way, \eg requiring costly test-time computations, 
or depth sensors, or multiple query frames. In contrast, our proposed method makes few special assumptions, and is fairly lightweight in training and testing.

Our pose regression network learns from only relative poses of training scenes. For inference, it builds a graph connecting the query image to training counterparts and uses a graph neural network (GNN) with image representations on nodes and image-pair representations on edges. By efficiently passing messages between them, both representation types are refined to produce a consistent camera pose estimate. We validate the effectiveness of our approach on both standard indoor (7-Scenes) and outdoor (Cambridge Landmarks) camera re-localization benchmarks.  Our relative pose regression method matches the accuracy of absolute pose regression networks, while retaining the relative-pose models' test-time speed and ability to generalize to non-training scenes.

\end{abstract}

\section{Introduction}
\begin{figure*}
    \centering
    \includegraphics[width=\textwidth]{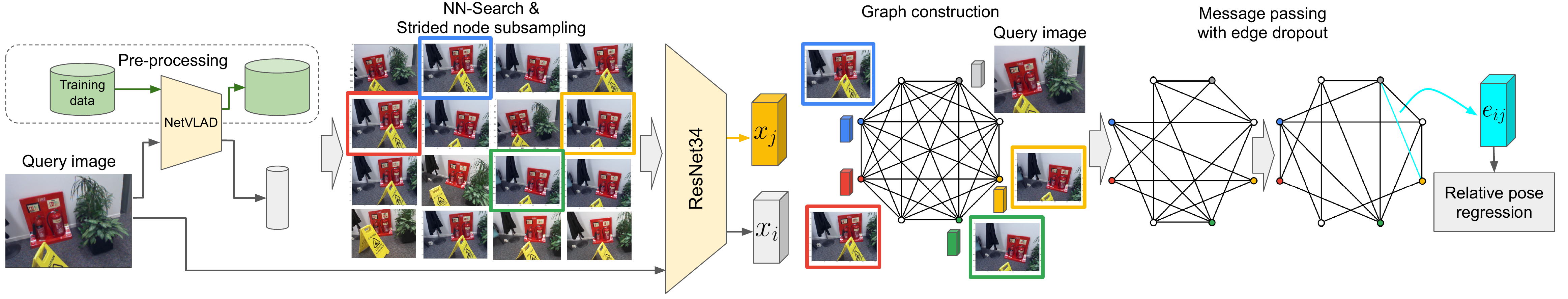}
    \caption{\textbf{Relative pose regression.} We combine the efficiency of image retrieval methods and the ability of graph neural networks to selectively and iteratively refine estimates to solve the challenging relative pose regression problem. Given a query image, we first find similar images to it using a differentiable image retrieval method~\cite{netvlad}. We preserve the diversity of neighbors by strided subsampling before building a fully connected GNN. Node representations $x_i$ are initialized from ResNet34 \cite{he2016deep}, and are combined using MLP-s into edge features $e_{ij}$. Finally, the relative pose regression layer maps the refined edge representations into relative poses between image pairs. Edge dropout is only applied at training time.}
    \label{fig:teaser}
\end{figure*}

Ideally, camera re-localization requires only minimal information when building a map or training a representation, and then accurately infers the position and orientation (6-DoF) of a query image at test time. 
Decidedly not ideal, approaches in the ``structure-based" category~\cite{sattler2019understanding} achieve state of the art accuracy at the expense of needing the camera intrinsics, and often expensive geometric optimization too. The structure refers to matching of $2D$ image feature points to $3D$ counterpart coordinates, followed by a PnP solve. 
So why look beyond structure-based methods when they score so well? Intrinsics are often not available or reliable, and geometric optimization is costly. Also, these methods work best for scenes with easy-to track feature points and large numbers of images~\cite{sattler2016efficient}, making sure each feature point has enough observations from different-enough viewpoints. They fail in fairly common situations, such as with texture-less scenes and distracting images with foliage or bricks~\cite{frahm2010building}, where the features 
are distributed unevenly and with potentially many false matches.

Pose-regression methods are an important growth area, attempting to overcome those limitations, and eventually aiming to produce models that generalize well to new scenes. We are particularly motivated by so-called ``model-less'' approaches, where training can proceed with just pairwise relative-pose estimates between training images, instead of the absolute-pose approaches that require a full 3D reconstruction of all footage to get started.

With this context, our overall contributions are that 
\begin{itemize}
    \item We are the first to apply GNNs successfully for relative pose re-localization with one query image. \cite{glnet} was a first attempt to use graph neural networks (GNNs) for the absolute camera pose regression setting, but that requires a chain of multiple query images to succeed. 
    \item Our novel differentiable camera pose regression model is also practical: Discounting structure-based methods because they require camera intrinsics, we achieve best-in-class rotation scores, and very competitive (2$^{nd}$ or 1$^{st}$ place) translation scores. 
    When trained and generalizing to re-localize in unseen scenes, we need a few minutes to be able to start re-localizing. 
    In contrast, structure-based methods such as~\cite{sattler2016efficient} take up to 10 hours, and many scene coordinate regression methods, \eg~\cite{Brachmann2018CVPR}, take $>$10 hours per scene (as confirmed with authors).
    \ifmyarxiv
    Also, our code will be online to avoid becoming one of the unreproducible baselines.  
    \else
    Also, our code, submitted here, will be online, to avoid becoming one of the unreproducible baselines.  
    \fi
\end{itemize}

\section{Related Work}
Approaches to re-localization have diverse pros and cons. We review the main categories of algorithms, especially featuring the expanding literature on learning-based methods. While many criteria are important, we highlight accuracy, what is needed when building/learning the map for a place, what ingredients are needed at test-time, and what coordinate representation is used.

\subsection{Structure-based Methods}
Traditionally, visual re-localization has been addressed by sparse feature matching to obtain 2D-3D correspondences, and recovering the camera pose through robust optimization. 
\emph{Direct approaches}~\cite{cao2014minimal,li2012worldwide,sattler2016efficient,svarm2016city,zeisl2015camera} match feature descriptors of the query image to descriptors associated with the 3D points of a Structure-from-Motion (SfM) model. 
For very large scenes, the SfM model comes with a significant memory footprint that can be prohibitive, \eg for mobile applications \cite{brachmann2020dsacstar,Camposeco2019CVPR,sattler2017large}.
\emph{Indirect approaches}~\cite{sattler2015hyperpoints, sattler2017large} first search for top-$k$ most similar images in the database, build a \emph{local} SfM model from the retrieved images, then proceed like direct approaches. 
While requiring less memory than direct methods, calculating SfM models on-the-fly is brittle and computationally demanding.
Instead of matching feature descriptors, \emph{scene coordinate regression methods}~\cite{7scenes} learn to directly predict correspondences between a query image and 3D scene space.
Scene coordinate regression works well in small-scale indoor environments \cite{Brachmann2017CVPR,Brachmann2016CVPR, Brachmann2018CVPR,Brachmann2019ICCVb,brachmann2020dsacstar,Cavallari20193DV,Cavallari2019TPAMI,Cavallari2017CVPR,Guzman2014CVPR,Massiceti2017ICRA,Valentin2015CVPR}, but is less accurate in larger environments \cite{Brachmann2019ICCVa}. 
Structure-based methods are thus far unchallenged in pose accuracy but, besides the computational demands of large-scale correspondence search, they require known camera intrinsics and a structured 3D representation of the scene, either from SfM \cite{cao2014minimal,sattler2016efficient,zeisl2015camera, li2012worldwide,svarm2016city,sattler2015hyperpoints, sattler2017large}, via implicit triangulation \cite{Brachmann2018CVPR, brachmann2020dsacstar} or by scanning with a depth sensor \cite{7scenes, Guzman2014CVPR, Valentin2015CVPR, Cavallari2017CVPR}.
Obtaining such a reconstruction can be challenging in the presence of scene ambiguities, lighting variations, or unfavorable imaging geometry~\cite{poselstm}.
Scanning with a depth sensor is an alternative but requires specialized hardware \cite{taira2018inloc} and, by itself does not provide image texture.

\subsection{Absolute Pose Regression}
Absolute pose regression fits a regressor -- normally a convolutional neural network (CNN) -- to directly predict the 6-DoF camera pose from an RGB image. 
\mbox{PoseNet~\cite{kendall2015posenet}} was the first attempt to show that CNNs can estimate poses under challenging image conditions such as underexposure and motion blur, where SIFT\cite{Lowe04IJCV}-based methods fail. 
Geo-PoseNet~\cite{kendall2017geometric} added geometrically inspired pose losses that significantly improve the orientation accuracy.
Bayes-PoseNet~\cite{posenet16} uses a Bayesian CNN to also obtain an estimate of the model’s re-localization uncertainty. 
\mbox{MLFBPPose~\cite{wang2019discriminative}} proposes a multi-layer factorized bi-linear pooling module for feature aggregation, while \mbox{PoseLSTM~\cite{poselstm}} uses an LSTM on top of a CNN to reduce the dimension of the visual feature vector and exploit feature correlation. 
Recently, AttLoc~\cite{wang2020atloc} proposed an attention mechanism on top of a CNN to exploit the feature correlation as well. 
AnchorPoint~\cite{saha2018improved} uses visible landmarks for visual localization. 
Anchor points are distributed uniformly across the environment, such that the network, when shown a query image, can predict the most relevant anchor points together with the relative pose \wrt those anchor points.   
In summary, absolute pose regression exhibits good re-localization accuracy, but has to be trained on the specific target scene (and excluding data from other scenes). This means one pays a high price for accuracy. This is aggravated by the fact that, inevitably, data is initially limited every time one moves to a new target scene~\cite{brachmann2020dsacstar,sattler2019understanding}.

\subsection{Sequence-based Methods}
Sequence-based methods use multiple query frames to estimate 6-DoF camera poses. 
They are generally trained with both absolute and relative pose losses.
VidLoc~\cite{vidloc} proposes a recurrent localization model.
Using short video clips instead of a single image as input, they estimate smooth pose trajectories and substantially reduce the localization error. However, the model is limited to pure camera translation.
MapNet~\cite{mapnet} relies on additional inputs such as visual odometry and GPS to learn camera re-localization. 
Geometric constraints due to these inputs offer additional loss terms that support training and, optionally, also inference.
LsG~\cite{lsg} exploits the spatio-temporal consistency in image sequences, by estimating relative poses between consecutive frames with convLSTM, which are then used to constrain the final pose graph.
Valada~\etal~\cite{valada2018incorporating} encode geometric and semantic knowledge about the scene into a pose regression network. 
They employ multi-task learning to exploit the correlations between semantics and 6-DoF absolute pose. 
Sequence-based approaches usually outperform single-image methods, but have fewer use cases. 

\subsection{Relative Pose Regression}
In contrast to the works described so far, relative pose regression does not require training specific to the test-scene, which greatly improves scalability.
Retrieval-based methods such as DenseVLAD~\cite{denseVLAD} or NetVLAD \cite{netvlad} find the most similar training image and assign its pose to the test image.
RelocNet~\cite{relocnet} learns image retrieval via nearest-neighbor matching and is trained with a continuous metric learning loss.
The query pose is predicted by fine-grained relative pose regression \wrt the retrieved database image.
In a similar way, NN-Net~\cite{nn-net} also retrieves similar images and then estimates relative poses.
Different from RelocNet, the location of the query image is first triangulated from relative translations, followed by robust fusion of rotation estimates.
CamNet~\cite{camnet} proposes a 3-stage approach where they first retrieve a database image, then retrieve new database images around their coarse relative pose estimate w.r.t. the first image, and finally regress the absolute pose using the absolute poses of the second set. 
Similarly, it is trained with a continuous metric learning loss.
Quite recent, EssNet~\cite{essnet} regresses an essential matrix between the query and a database image, retrieved by a Siamese neural network.
Relative pose regression does not yet achieve the accuracy of correspondence-based methods, except for CamNet, where the official code is unable to reproduce the reported accuracy. Nonetheless, pose regression is differentiable and efficient at inference time, so constitutes an attractive drop-in component to build larger learning-based vision systems such as differentiable SLAM~\cite{ummenhofer2017demon}.

\subsection{Graph Neural Networks}
Graph neural networks (GNNs) are effective where rich relational structure exists in the data~\cite{battaglia2018relational}. 
They naturally adapt to non-regular data layouts, like molecules~\cite{duvenaud2015convolutional}, citation networks~\cite{gcn_welling}, or point clouds~\cite{few_shot_2020_CVPR}. 
GNNs can be grouped into two main categories, \emph{spectral}~\cite{hamilton2017inductive,gcn_welling,defferrard2016convolutional} and \emph{spatial}~\cite{niepert2016learning,micheli2009neural,gin} methods. 
The former are based on the eigen-decomposition of the Graph Laplacian, whereas the latter operate on local neighborhoods of nodes. We use a spatial GNN approach. 
Many GNN variants have been developed~\cite{gcn_welling,hamilton2017inductive, gat, gin} to classify either individual nodes or entire graphs.
Graph Attention Networks (GAT)~\cite{gat} combine GNNs with a learnable attention mechanism that enables them to optimally weight information coming from different neighbors. In our method, we include an attention mechanism that differs from GAT, as explained in Sec.~\ref{sec:graph_MP}.

GNNs have been applied to a variety of computer vision tasks, \eg few-shot learning~\cite{graph_few_shot, few_shot_2020_CVPR}, human action recognition~\cite{action_2020_CVPR}, motion prediction~\cite{motion_2020_CVPR}, and feature matching~\cite{superglue_2020_CVPR}.
Recently, GL-Net~\cite{glnet} has employed them for camera pose estimation. 
The GNN replaces an RNN, to allow the exchange of information between non-consecutive frames of a video clip. 
GL-Net is perhaps the closest to our approach, in that it also employs GNNs for camera re-localization.
We are, however, rather different in many aspects: (i) GL-Net is a multi-frame approach, whereas our method requires only a single query image. (ii) GL-Net requires absolute poses for training, while our method is designed for supervision with only relative poses. Thus, we can generalize to an unseen scene, while GL-Net can not. 
Note that GL-Net employs relative pose loss as a regularizer to their absolute pose regression task, which does not make it a relative pose approach. (iii) The network design is different: our edge features and a non-local attention mechanism play an important role during graph message passing, whereas GL-Net focuses on multi-scale modeling. 
GL-Net's attention mechanism is quite different to ours. It gives more weight to messages coming from more similar frames. This means that it demotes different viewpoints where our attention individually adapts messages between viewpoints based on their non-local content.

\section{Method}
We now describe in detail the proposed method, see \Cref{fig:teaser}. It consists of three main parts: visual encoding, image retrieval, and graph message passing. All the different parts are differentiable, so including image retrieval, our method can be trained end-to-end.

\subsection{Visual Encoding}
The visual encoder is a convolutional neural network that extracts features from a single RGB image. Following recent camera re-localization works~\cite{mapnet,wang2020atloc,glnet}, we use the ResNet34 \cite{he2016deep} architecture for the encoder. It takes a single RGB image, $I\!\in\!\mathbb{R}^{3\times H \times W}$ and generates a 1-D feature vector, $x\!\in\!\mathbb{R}^{C}$ with size of $C$. We set $C\!=\!2048$ which serves a good trade off for accuracy \vs computation. More formally, the visual encoding is
\begin{equation}
    x = f_\text{CNN}(I) \ .
\end{equation}
The weights of $f_\text{CNN}$ are initialized with the ResNet34, trained on ImageNet~\cite{imagenet}.

\subsection{Image Retrieval \& Graph Construction}
\label{sec:graph_construc}
To construct a graph for both training and inference, we employ \mbox{NetVLAD~\cite{netvlad}} for image retrieval. It contains a generalized VLAD layer, inspired by the \textit{Vector of Locally Aggregated Descriptors} image encoding~\cite{vlad}. This too is differentiable. 

First, we create a database for NetVLAD with all the training images in the dataset.
Thereafter, in the training stage, for each anchor training image we retrieve the $(N\!-\!1)K$ most similar images in the embedding feature space. 
We then randomly choose an offset $0\!\leq\!k\!<\!K$ and sub-sample the nearest neighbors with regular intervals yielding every $K$-th image in the ordered list of similar images.
We construct a fully connected graph from the $N\!-\!1$ retrieved images and the anchor image. 
Thus each graph consists of $N$ nodes and $N(N\!-\!1)/2$ bi-directional edges. 
In our experiments we set $N\!=\!8$, and $K\!=\!5$ for 7-Scenes and $K\!=\!3$ for Cambridge Landmarks. 
This method allows us to favorably balance similarity and diversity of images from the training set, see \Cref{t:ablation} for simpler graph building strategies.

In the test stage, following a similar procedure, $(N\!-\!1)$ images are retrieved from the database of training images.
We then construct a query graph with a single test query image and $N\!-\!1$ training images. 

\textbf{Graph construction}: 
A graph is defined as $G\!=\!\{V,E\}$ where $V\!=\!\{v_i\}$ represents the set of $N$ nodes, and $E\!=\!\{v_{ij}\}$ the set of bi-directional edges connecting the nodes. Our graphs are fully connected: $E\!=\!\{v_{ij}\!=\!1 | i\!\neq\!j\}$.
Each node has a feature representation denoted as $x_i$ and each edge has a feature representation denoted as $e_{ij}$.
Node features $x_i$ are initialized with feature vectors from the visual encoder.
Edge features are initialized by concatenating features of their endpoints and projecting them with a learned linear mapping $f_\text{proj}$, so 
\begin{equation}
    e_{ij}:= f_\text{proj}\big([x_i,x_j]\big) \ ,
\end{equation}
where $f_\text{proj}$ is a single fully-connected layer with ReLU activations and $[.,.]$ denotes \emph{concatenation} (throughout paper).

\begin{figure*}[ht]
    \centering
    \includegraphics[trim=0 12.4 0 12.4,clip,width=.92\textwidth]{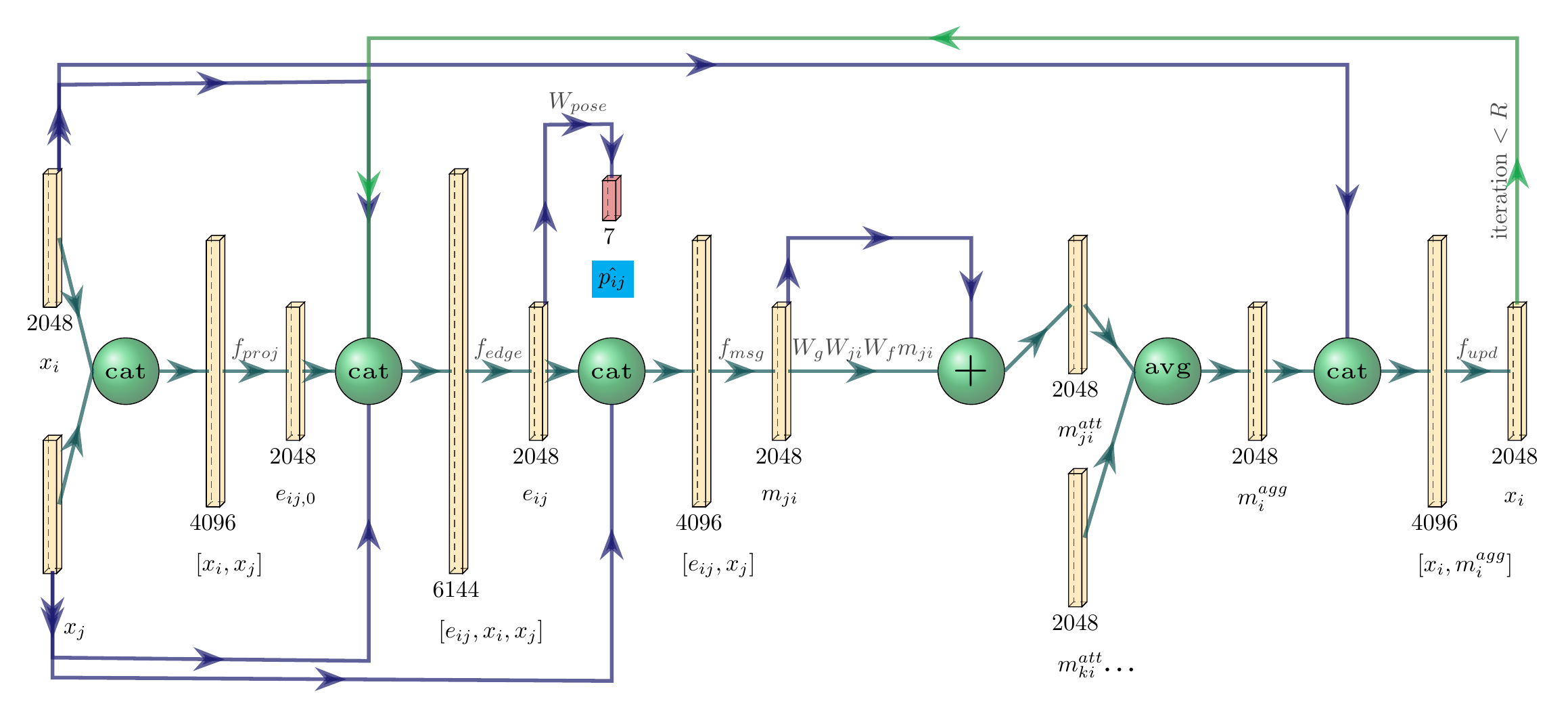}
    \caption{\textbf{Message passing.} Edge features $e_{ij}$ representing information about an image pair are initialized from node features $x_i$ and $x_j$, and iteratively updated during message passing. Afterwards, we regress the relative pose $\hat{p_{ij}}$ between the two images and apply the relative pose loss. We set $R=2$ in our experiments, equivalent to a 2-layer GNN with shared weights across iterations.}
    \label{fig:messagepassing}
\end{figure*}

\subsection{Graph Message Passing}
\label{sec:graph_MP}
Once the graph has been built, we exchange information between different images $v_i$ via message passing, see \Cref{fig:messagepassing}. First, edge features are updated in an auto-regressive manner, using the features of the edge's endpoints,
\begin{equation}
    e_{ij}:= f_\text{edge}\big([e_{ij}, x_i,x_j]\big) \ ,
\end{equation}
where $f_\text{edge}$ is a 2-layer MLP with ReLU activations.
After the edge features have been updated, messages are generated to each node (resp.\ image) $v_i$ from all its neighbors $v_j$, taking into account the edge connecting the two vertices,
\begin{equation}
    m(v_j\!\rightarrow\!v_i) = m_{ji}=f_\text{msg}\big([e_{ij}, x_j]\big) \ ,
\end{equation}
where $f_\text{msg}$ is another 2-layer MLP with ReLU activations.

Intuitively, \emph{what information is important and should be exchanged along the edges varies throughout the graph}. Some image pairs might share a salient geometric object that is well-suited for pose estimation, whereas in another image pair, one of the two views might see a moving object or a cast shadow that should be ignored.
Consequently, the messages sent between adjacent images should be individually adapted based on their content, as encoded in the features $x_j$.
To that end we extend the message passing with a self-attention mechanism. We use the \emph{non-local attention} scheme proposed by \cite{wang2018non}. That attention method is generic and has been used for different applications ranging from keypoint detection to video classification. It has also proved effective in the context of camera localization \cite{wang2020atloc}.
Our attention mechanism interprets the message $m_{ji}$ as an (unordered) sequence of 1-dimensional features, and 
computes the response at a given element as a weighted sum over \emph{all} elements.
In particular, the message is compressed to lower dimension with a learned mapping $W_f$, multiplied with an attention map $W_{ji}$, and upsampled back to its original size with another learned mapping $W_g$.
The attention map is the outer product of two learned, low-dimensional embeddings of the message.
 Formally, 
\begin{align}
W_{ji}&=\mathrm{softmax}(W_\theta m_{ji} m_{ji}^\top  W_\phi^\top ) \ , \\
a_{ji}&=W_gW_{ji}W_fm_{ji} \ , \\
m_{ji}^\text{att}&:=m_{ji}+a_{ji} \ ,
\label{eq:attention}
\end{align}
with $W_g\!\in\!\mathbb{R}^{C \times \frac{C}{n}}$, and $W_f\!\in\!\mathbb{R}^{\frac{C}{n} \times C}$ as learnable up-/down-sampling operators by a factor $n$, that are optimized by back-propagation. We set $n\!=\!8$ as in \cite{wang2020atloc}. $W_{\theta}$, $W_{\phi}$ are learnable parameters of the attention mechanism, also found via back-propagation.

Once messages from all neighbors have been received, we compute the aggregated message, $m_{i}^\text{agg}$ for node $x_i$ by taking their mean. We found empirically that \textit{mean} aggregation is more effective than \textit{max} or \textit{sum} aggregation, so 
\begin{equation}
    m_{i}^\text{agg} := \frac{1}{N_i} \sum_{e_{ij}\in \varepsilon}^{} m_{ji}^\text{att} \ .
\end{equation}
Finally, we update the node features by concatenating the current node feature and the aggregated message as 
\begin{equation}
    x_i := f_\text{upd}\big([x_i,  m_{i}^\text{agg}]\big) \ ,
\end{equation}
where $f_\text{upd}$ is yet another 2-layer MLP.
Message passing over the entire graph is repeated $R$ times, with weights shared across iterations.
More iterations improve the accuracy, with quickly diminishing returns and increasing training time.
We found $R\!=\!2$ to be a good trade-off.

\subsection{Learning Relative Poses}
As final step, our model regresses the 6-DoF relative camera translation and orientation from the updated edge features in the graph.
It maps edge features $e_{ij}$ in the graph to relative translation $\hat{t}_{ij} \in \mathbb{R}^{3}$ and additive offset to the logarithm of the unit quaternion $log(\hat{q})_{ij} \in \mathbb{R}^3$
with a linear projection, so
\begin{equation}
    \hat{p_{ij}} = [\hat{t}_{ij},{\log(\hat{q})}_{ij}] = W_\text{pose}e_{ij} + b \ .
\end{equation}
Similar to \mbox{MapNet~\cite{mapnet}}, we parameterize camera orientation as the logarithm of a unit quaternion, equivalent to the axis-angle representation up to scale \cite{HartleyZisserman}, as it does not require any additional constraints to represent a valid rotation. 
Logarithmic mapping is done using
\begin{equation}
    \log(q) =      
    \begin{cases}
      \frac{v}{\|v\|}\cos^{-1}(u) & \text{if $\|v\| \neq 0$} \\
      0 & \text{otherwise.}
    \end{cases}  
\end{equation}
Here, $u$ and $v$ denote the real part and the imaginary part of a unit quaternion, respectively. The logarithmic form $w = \log(q)$ can be converted back to a unit quaternion by the exponential mapping, $e^{w}=[\cos(\|w\|), \frac{w}{\|w\|}\sin(\|w\|)]$.

To learn the parameters of the model, our method minimizes the loss between two image pairs in the graph.
The objective function is defined as
\begin{equation}
    L = \frac{1}{N_e} \sum_{v_{ij} \in E} d(\hat{p}_{ij},p_{ij}) \ .
\end{equation}
Here, $N_e$ is the number of edges, $d(\cdot)$ is the distance function between the predicted relative camera pose $\hat{p}_{ij}$ and the ground truth relative camera pose $p_{ij}$.
We use the distance proposed by \cite{kendall2017geometric}, and adapted as
\begin{equation}
    d(\hat{p}_{ij},p_{ij})=  |\hat{t}_{ij}- t_{ij}| e^{-\beta}\!+ \beta + |\log(\hat{q})_{ij} - \log(q)_{ij}| e^{-\gamma}\!+ \gamma \ ,
\end{equation}
where $\beta$ and $\gamma$ are the weights that balance the translation loss and rotation loss, and $\log(q)_{ij} = \log(q)_{j} - \log(q)_{i}$. Both $\beta$ and $\gamma$ are learned during training with initialization $\beta_0$ and $\gamma_0$. In the experiments, we initialize these parameters to $0$ and $-3$, respectively.

\subsection{Inference}
In order to estimate the absolute pose of a given test query image we need access to training images and their absolute poses. To do that, as detailed in \ref{sec:graph_construc}, first the query graph is constructed with a single query image and $N\!-\!1$ training images. 
Then, our method infers the relative poses on each edge.
We can compute the absolute pose of the query image by choosing one training image and using the estimated relative pose of the edge between the chosen training and query image.
$N\!-\!1$ such estimates can be computed for the absolute pose of the given query image.
Robust geometric averaging can be performed as post-processing, \eg using Weiszfeld's algorithm for translation and \cite{gramkow2001averaging}'s for quaternions.
In our experiments, we observed that improvements compared to using the nearest neighbor from image retrieval are negligible.
Therefore, we choose the most similar training image from the image retrieval step to compute the absolute pose of the query image.
See supplementary materials for comparisons.

\section{Experiments}
\ifmyarxiv
\begin{table*}[!ht]
\centering
\caption[Comparison to state-of-the-art methods: 7-Scenes.]{{\bf Comparison to state-of-the-art methods: 7-Scenes.} 
Median translation and rotation error. Methods marked with a * have scene-specific representations and do not generalize to unseen scenes. For our results of models trained on one scene only, please see the supplementary material.
Results which we could not reproduce despite our best effort are marked with ''?''.
We also added the per-scene results provided by the authors of \mbox{NC-EssNet~\cite{essnet}} (which are reproducible with the published code, \cf ``EssNet reprod.\ \emph{Stairs}'' and ``NC-EssNet reprod.\ \emph{Chess}'').
}
\label{t:comparison}
\small
\setlength\tabcolsep{2pt} 
\begin{tabular}{clccccccccc}
\midrule
 & & $\!\!\!\!\!\!\!$\# test frames & Chess & Fire & Heads & Office & Pumpkin & Kitchen & Stairs & Avg. \\
\toprule
  & \mbox{DSAC\textsuperscript{*}~\cite{brachmann2020dsacstar}}*
  & 1
  & \tableentry{0.02}{1.1} 
  & \tableentry{0.02}{1.2} 
  & \tableentry{0.01}{1.8}
  & \tableentry{0.03}{1.2}
  & \tableentry{0.04}{1.4}
  & \tableentry{0.03}{1.7}
  & \tableentry{0.04}{1.4}
  & \tableentry{0.03}{1.4}
  \\
  \noalign{\hrule height 1pt}
  \multirow{4}{*}{\STAB{\rotatebox[origin=c]{90}{Seq. based}}}
  & \mbox{VidLoc~\cite{vidloc}}*
  & 200
  & \tableentrytra{0.18} 
  & \tableentrytra{0.26} 
  & \tableentrytra{0.14} 
  & \tableentrytra{0.26} 
  & \tableentrytra{0.36} 
  & \tableentrytra{0.31} 
  & \tableentrytra{0.26} 
  & \tableentrytra{0.25} 
  \\
  & \mbox{LsG~\cite{lsg}}*
  & 7
  & \tableentry{0.09}{3.3} 
  & \tableentry{0.26}{10.9} 
  & \tableentry{0.17}{12.7} 
  & \tableentry{0.18}{5.5} 
  & \tableentry{0.20}{3.7} 
  & \tableentry{0.23}{4.9} 
  & \tableentry{0.23}{11.3} 
  & \tableentry{0.19}{7.5} 
  \\
  & \mbox{MapNet~\cite{mapnet}}*
  & 3
  & \tableentry{0.08}{3.3} 
  & \tableentry{0.27}{11.7} 
  & \tableentry{0.18}{13.3} 
  & \tableentry{0.17}{5.2} 
  & \tableentry{0.22}{4.0} 
  & \tableentry{0.23}{4.9} 
  & \tableentry{0.30}{12.1} 
  & \tableentry{0.21}{7.8} 
  \\
  & GL-Net~\cite{glnet}*\textsubscript{?}
  & 8
  & \tableentry{{0.08}}{{2.8}} 
  & \tableentry{0.26}{{8.9}} 
  & \tableentry{0.17}{11.4} 
  & \tableentry{0.18}{{5.1}} 
  & \tableentry{{0.15}}{{2.8}} 
  & \tableentry{0.25}{{4.5}} 
  & \tableentry{0.23}{8.8} 
  & \tableentry{0.19}{{6.3}} 
  \\
\midrule
\midrule
  \multirow{11}{*}{\STAB{\rotatebox[origin=c]{90}{Image based APR}}}
  & PoseNet~\cite{kendall2015posenet}*
  & 1
  & \tableentry{0.32}{6.6} 
  & \tableentry{0.47}{14.0} 
  & \tableentry{0.30}{12.2} 
  & \tableentry{0.48}{7.2} 
  & \tableentry{0.49}{8.1} 
  & \tableentry{0.58}{8.3} 
  & \tableentry{0.48}{13.1} 
  & \tableentry{0.45}{9.9} 
  \\
  & Bayesian PoseNet \cite{posenet16}*
  & 1
  & \tableentry{0.37}{7.2} 
  & \tableentry{0.43}{13.7} 
  & \tableentry{0.31}{12.0} 
  & \tableentry{0.48}{8.0} 
  & \tableentry{0.61}{7.1} 
  & \tableentry{0.58}{7.5} 
  & \tableentry{0.48}{13.1} 
  & \tableentry{0.47}{9.8} 
  \\
  & Geometric PoseNet \cite{kendall2017geometric}*
  & 1
  & \tableentry{0.13}{4.5} 
  & \tableentry{0.27}{11.3} 
  & \tableentry{0.17}{13.0} 
  & \tableentry{0.19}{5.6} 
  & \tableentry{0.26}{4.8} 
  & \tableentry{0.23}{5.4} 
  & \tableentry{0.35}{12.4} 
  & \tableentry{0.23}{8.1} 
  \\
  & MLFBPPose \cite{wang2019discriminative}*
  & 1
  & \tableentry{0.12}{5.8} 
  & \tableentry{0.26}{12.0} 
  & \tableentry{0.14}{13.5} 
  & \tableentry{0.18}{8.2} 
  & \tableentry{0.21}{7.1} 
  & \tableentry{0.22}{8.1} 
  & \tableentry{0.26}{13.6} 
  & \tableentry{0.20}{9.8} 
  \\
  & Hourglass \cite{melekhov2017image}*
  & 1
  & \tableentry{0.15}{6.2}  
  & \tableentry{0.27}{10.8} 
  & \tableentry{0.19}{11.6} 
  & \tableentry{0.21}{8.5}  
  & \tableentry{0.25}{7.0}  
  & \tableentry{0.27}{10.2} 
  & \tableentry{0.29}{12.5} 
  & \tableentry{0.23}{9.5}  
  \\
  & LSTM-Pose \cite{poselstm}*
  & 1
  & \tableentry{0.24}{5.8}  
  & \tableentry{0.34}{11.9} 
  & \tableentry{0.21}{13.7} 
  & \tableentry{0.30}{8.1}  
  & \tableentry{0.33}{7.0}  
  & \tableentry{0.37}{8.8}  
  & \tableentry{0.40}{13.7} 
  & \tableentry{0.31}{9.9}  
  \\
  & BranchNet \cite{wu2017delving}*
  & 1
  & \tableentry{0.18}{5.2}  
  & \tableentry{0.34}{\ \underline{9.0}}  
  & \tableentry{0.20}{14.2} 
  & \tableentry{0.30}{7.1}  
  & \tableentry{0.27}{5.1}  
  & \tableentry{0.33}{7.4}  
  & \tableentry{0.38}{10.3} 
  & \tableentry{0.29}{8.3}  
  \\
  & ANNet \cite{bui2019adversarial}*
  & 1
  & \tableentry{0.12}{4.3}  
  & \tableentry{0.27}{11.6} 
  & \tableentry{0.16}{12.4} 
  & \tableentry{0.19}{6.8}  
  & \tableentry{0.21}{5.2}  
  & \tableentry{0.25}{6.0}  
  & \tableentry{0.28}{8.4}  
  & \tableentry{0.21}{7.9}  
  \\
  & GPoseNet \cite{cai2019hybrid}*
  & 1
  & \tableentry{0.20}{7.1}  
  & \tableentry{0.38}{12.3} 
  & \tableentry{0.21}{13.8} 
  & \tableentry{0.28}{8.8}  
  & \tableentry{0.37}{6.9}  
  & \tableentry{0.35}{8.2}  
  & \tableentry{0.37}{12.5} 
  & \tableentry{0.31}{10.0}  
  \\
  & AttLoc \cite{wang2020atloc}*
  & 1
  & \tableentry{0.10}{4.1} 
  & \tableentry{0.25}{11.4} 
  & \tableentry{0.16}{11.8} 
  & \tableentry{0.17}{\underline{5.3}} 
  & \tableentry{0.21}{4.4} 
  & \tableentry{0.23}{5.4} 
  & \tableentry{0.26}{10.5} 
  & \tableentry{0.20}{7.6} 
  \\ 
  & \mbox{AnchorPoint~\cite{saha2018improved}*}\textsubscript{?}
  & 1
  & \tableentry{\textbf{0.06}}{\underline{3.9}} 
  & \tableentry{\textbf{0.16}}{11.1} 
  & \tableentry{\textbf{0.09}}{11.2} 
  & \tableentry{\textbf{0.11}}{5.4} 
  & \tableentry{\textbf{0.14}}{\underline{3.6}} 
  & \tableentry{\textbf{0.13}}{5.3} 
  & \tableentry{\textbf{0.21}}{11.9} 
  & \tableentry{\underline{0.13}}{7.5} 
  \\ 
\midrule  
  \multirow{2}{*}{\STAB{\rotatebox[origin=c]{90}{IR}}}
  & \mbox{DenseVLAD~\cite{denseVLAD}}
  & 1
  & \tableentry{0.21}{12.5} 
  & \tableentry{0.33}{13.8} 
  & \tableentry{0.15}{14.9} 
  & \tableentry{0.28}{11.2} 
  & \tableentry{0.31}{11.2} 
  & \tableentry{0.30}{11.3} 
  & \tableentry{0.25}{12.3} 
  & \tableentry{0.26}{12.5} 
  \\
  & \mbox{DenseVLAD+Inter~\cite{sattler2019understanding}}
  & 1
  & \tableentry{0.18}{10.0} 
  & \tableentry{0.33}{12.4} 
  & \tableentry{0.14}{14.3} 
  & \tableentry{0.25}{10.1} 
  & \tableentry{0.26}{9.4} 
  & \tableentry{0.27}{11.1} 
  & \tableentry{0.24}{14.7} 
  & \tableentry{0.24}{11.7} 
  \\
\midrule 
  \multirow{6}{*}{\STAB{\rotatebox[origin=c]{90}{RPR}}}
  & \mbox{NN-Net~\cite{nn-net}}
  & 1
  & \tableentry{0.13}{6.5} 
  & \tableentry{0.26}{12.7} 
  & \tableentry{0.14}{12.3} 
  & \tableentry{0.21}{7.4} 
  & \tableentry{0.24}{6.4} 
  & \tableentry{0.24}{8.0} 
  & \tableentry{0.27}{11.8} 
  & \tableentry{0.21}{9.3} 
  \\
  & \mbox{RelocNet~\cite{relocnet}}
  & 1
  & \tableentry{0.12}{4.1} 
  & \tableentry{0.26}{10.4} 
  & \tableentry{0.14}{\underline{10.5}} 
  & \tableentry{0.18}{\underline{5.3}} 
  & \tableentry{0.26}{4.2} 
  & \tableentry{0.23}{\underline{5.1}} 
  & \tableentry{0.28}{\underline{7.5}} 
  & \tableentry{0.21}{6.7} 
  \\
    & \mbox{EssNet~\cite{essnet}}
  & 1
  & \tableentry{0.13}{5.1} 
  & \tableentry{0.27}{10.1} 
  & \tableentry{0.15}{9.9} 
  & \tableentry{0.21}{6.9} 
  & \tableentry{0.22}{6.1} 
  & \tableentry{0.23}{6.9} 
  & \tableentry{0.32}{11.2} 
  & {\tableentry{0.22}{8.0}} 
  \\
      & \mbox{EssNet~\cite{essnet} reprod.}
  & 1
  & - 
  & - 
  & - 
  & - 
  & - 
  & - 
  & \tableentry{0.32}{9.8} 
  & - 
  \\
    & \mbox{NC-EssNet~\cite{essnet}}
  & 1
  & \tableentry{0.12}{5.6} 
  & \tableentry{0.26}{9.6} 
  & \tableentry{0.14}{10.7} 
  & \tableentry{0.20 }{6.7} 
  & \tableentry{0.22}{5.7} 
  & \tableentry{0.22}{6.3} 
  & \tableentry{0.31}{7.9} 
  & {\tableentry{0.21}{7.5}} 
  \\
     & \mbox{NC-EssNet~\cite{essnet} reprod.}
  & 1
  & \tableentry{0.13}{5.5} 
  & - 
  & - 
  & - 
  & - 
  & - 
  & - 
  & - 
  \\
  & CamNet \cite{camnet}\textsubscript{?} 
  & 1
  & - 
  & - 
  & - 
  & - 
  & - 
  & - 
  & - 
  & \tableentry{\textbf{0.05}}{\textbf{1.8}} 
  \\ 
\midrule
  & Ours 
  & 1
  & \tableentry{\underline{0.08}}{\textbf{2.7}} 
  & \tableentry{\underline{0.21}}{\textbf{7.5}} 
  & \tableentry{\underline{0.13}}{\textbf{8.7}} 
  & \tableentry{\underline{0.15}}{\textbf{4.1}} 
  & \tableentry{\underline{0.15}}{\textbf{3.5}} 
  & \tableentry{\underline{0.19}}{\textbf{3.7}} 
  & \tableentry{\underline{0.22}}{\textbf{6.5}} 
  & \tableentry{0.16}{\underline{5.2}} 
  \\
\bottomrule
\end{tabular}
\end{table*}

\else
\input{041_table_comp}
\fi
To validate the effectiveness of our approach we conduct multiple experiments on both widely used indoor and outdoor re-localization benchmarks.
\textbf{The 7-Scenes dataset}~\cite{7scenes} is a collection of tracked RGB-D camera frames containing scenes with highly repetitive structures (\eg \emph{Chess} and \emph{Stairs}), and plenty of texture-less regions (\eg the \emph{Fire} and \emph{Pumpkin} scenes), therefore it is very a challenging dataset for re-localization.
\textbf{The Cambridge Landmarks}~\cite{kendall2015posenet} is an urban environment dataset collected with a mobile phone RGB camera. It contains moving objects such as pedestrians and vehicles, and different lighting and weather conditions. Train and test images are taken from distinct walking paths rendering the re-localization task even more challenging.
\ifmyarxiv
\begin{table*}[t]
\centering
\caption{{\bf Generalization to an unseen environment.} All methods were trained on the training sets of six scenes, and evaluated on the test set of the remaining scene. We report median translation and rotation errors. 
The authors of \protect{\cite{nn-net}} provided us with the data files describing their dataset splits for the unpublished cases, we used their published training and evaluation code to reproduce their results. Training on the \emph{Heads} and \emph{Kitchen} scenes did not converge$^{\dagger}$. For them, we use the best models before divergence. Results for \cite{essnet} are produced with their official code. (NC-)EssNet generalizes poorly in this evaluation scenario. Overall, our method produces more reliable results than its competitors.
}\label{t:sixone}
\small
\setlength\tabcolsep{1pt} 
\begin{tabular}{lcccccccc|ccc}
\midrule
 & Chess & Fire & Heads & Office & Pumpkin & Kitchen & Stairs &&
 & \begin{tabular}{@{}c@{}} Avg. (3 scenes) \end{tabular}  
 & Avg.
 \\
\toprule
    \mbox{NN-Net~\cite{nn-net}} published
  & \tableentry{\textbf{0.27}}{13.1} 
  & - 
  & \tableentry{0.23}{15.0} 
  & -
  & -
  & \tableentry{\textbf{0.36}}{12.6} 
  & -
  &&& \tableentry{\textbf{0.29}}{13.6} 
  & - 
\\
\midrule
    \mbox{NN-Net~\cite{nn-net}} reprod.
  & \tableentry{0.33}{\textbf{11.5}} 
  & \tableentry{11.68}{\textbf{13.8}} 
  & 0.30$^{\dagger}$, 15.5\textdegree$^{\dagger}$ 
  & \tableentry{1.34}{\textbf{10.9}} 
  & \tableentry{\textbf{0.41}}{12.8} 
  & 1.68$^{\dagger}$, 12.9\textdegree$^{\dagger}$ 
  & \tableentry{0.44}{\textbf{13.6}} 
  &&
  & \tableentry{0.77}{\textbf{13.2}} 
  & \tableentry{2.31}{\textbf{13.0}} 
\\
\mbox{EssNet~\cite{essnet}} reprod.
& \tableentry{0.73}{37.6} 
& \tableentry{0.89}{67.6} 
& \tableentry{0.62}{28.5} 
& \tableentry{0.84}{36.3} 
& \tableentry{1.06}{33.3} 
& \tableentry{0.91}{36.1} 
& \tableentry{1.19}{42.1} 
&&
& \tableentry{0.75}{34.1} 
& \tableentry{0.89}{40.2} 
\\
\mbox{NCEssNet~\cite{essnet}} reprod.
& \tableentry{0.62}{24.2} 
& \tableentry{0.75}{23.7} 
& \tableentry{0.44}{25.6} 
& \tableentry{0.88}{28.0} 
& \tableentry{1.02}{24.5} 
& \tableentry{0.77}{20.8} 
& \tableentry{1.25}{36.5} 
&&
& \tableentry{0.61}{23.5} 
& \tableentry{0.82}{26.2} 
\\
\midrule
  Ours 
  & \tableentry{{0.29}}{12.8}
  & \tableentry{\ \textbf{0.45}}{15.7}
  & \tableentry{\textbf{0.19}}{\textbf{14.7}}
  & \tableentry{\textbf{0.42}}{12.5}
  & \tableentry{0.44}{\textbf{11.7}}
  & \tableentry{0.42}{{12.4}}
  & \tableentry{\textbf{0.35}}{15.5}
  &&
  & \tableentry{{0.30}}{13.3} 
  & \tableentry{\textbf{0.37}}{13.6} 
  \\
\bottomrule
\end{tabular}
\end{table*}
\else
\input{042_table_six_one}
\fi
\subsection{Comparison with Leading Baselines}\label{sec:comparison}
We compare our model to a multitude of recent methods for camera re-localization. 
We group these methods into four main categories: \emph{(i)} Sequence-based Absolute Pose Regression (APR), \emph{(ii)} Image-based APR, \emph{(iii)} Image Retrieval (IR), and \emph{(iv)} Relative Pose Regression (RPR).
Our method belongs in the $4^{th}$ category. There, the main competitors are \mbox{NN-Net~\cite{nn-net}}, \mbox{RelocNet~\cite{relocnet}} and \mbox{(NC-)EssNet \cite{essnet}}, followed by IR methods like the one used within our approach. 
See \Cref{t:comparison,t:cambridge} for \mbox{7-Scenes} and \mbox{Cambridge} Landmarks dataset results, respectively. We separate DSAC$^*$~\cite{brachmann2020dsacstar}, as the current leading structure-based approach. This is just for reference, since those methods have access to camera intrinsics and additional geometric optimization, giving them unassailable accuracy compared to more general-purpose models like ours. Among Sequence-based methods (i), GL-Net~\cite{glnet} has the best reported results, which are slightly worse than ours, despite using eight query images. Among single-image APR methods (ii), only AnchorPoint~\cite{saha2018improved} competes with us on translation (only), on the basis of their published scores. However, please see Sec~\ref{sec:Repro}. In our own category RPR (iv), only CamNet~\cite{camnet} reported a superior overall average, but please refer to Sec.~\ref{sec:Repro} again. For Cambridge, the results are similar: among models for re-localizing using a single color input image, only AnchorPoint's~\cite{saha2018improved} published accuracy rivals our own.

\subsubsection{Reproducibility Challenges}\label{sec:Repro}

We invested significant effort to reproduce the scores of \mbox{GL-Net \cite{glnet}}, \mbox{AnchorPoint \cite{saha2018improved}}, and \mbox{CamNet~\cite{camnet}}. CamNet is particularly important to us, because it published a leading translation and rotation Average Score for 7-Scenes, but not constituent scores of their image based version for each scene, and no scores for Cambridge. After initial contact with the authors about inferior scores from their GitHub version, no further information was available. Others logged the same reproducibility issues ({\small \href{https://github.com/dingmyu/CamNet/issues/7}{github.com/dingmyu/CamNet/issues/7}}), and confirmed unresponsiveness in our DM's.  
\mbox{AnchorPoint~\cite{saha2018improved}} has published code \cite{anchorPointGithub} but only predicts 4-DoF outputs, so the median translation error reported is a 2D distance, and does not come near their published scores. We list both sets of Cambridge scores for completeness. Usefully, the main contributions in their approach are orthogonal to our method's, suggesting an avenue for future work. While our scores are only best-in-class among easily-reproducible methods, we posit that is still beneficial to the community, just like methods that come second to commercial-but-secret recipes.

\ifmyarxiv
\begin{table*}[!ht]
\centering
\caption{{\bf Comparison to state-of-the-art methods: Cambridge Landmarks.} 
Median translation and rotation error. Methods marked with a * have scene-specific representations and do not generalize to unseen scenes. ''?'' indicates results we could not reproduce despite our best effort. The authors of \cite{essnet} provided us with their detailed results.}\label{t:cambridge}
\small
\setlength\tabcolsep{3pt} 
\begin{tabular}{llcccccccc}
\midrule
& & \# test frames & College & Hospital & Shop & Church & Court 
 & \begin{tabular}{@{}c@{}}  Avg. (4) \end{tabular} & \begin{tabular}{@{}c@{}}  Avg. (5) \end{tabular} \\
\toprule
  & \mbox{DSAC\textsuperscript{*}~\cite{brachmann2020dsacstar}}*
  & 1
  & \tableentry{0.18}{0.3} 
  & \tableentry{0.21}{0.4} 
  & \tableentry{0.05}{0.3} 
  & \tableentry{0.15}{0.5} 
  & \tableentry{0.34}{0.2} 
  & \tableentry{0.15}{0.4} 
  & \tableentry{0.19}{0.3} 
  \\
  \noalign{\hrule height 1pt}
\multirow{2}{*}{\STAB{\rotatebox[origin=c]{90}{Seq.}}}
  & \mbox{MapNet~\cite{mapnet}}*
  & 3
  & \tableentry{1.08}{1.9} 
  & \tableentry{1.94}{3.9} 
  & \tableentry{1.49}{4.2} 
  & \tableentry{2.00}{4.5} 
  & \tableentry{7.85}{3.8} 
  & \tableentry{1.63}{3.6} 
  & \tableentry{2.87}{3.7} 
  \\
  & \mbox{GL-Net~\cite{glnet}*}\textsubscript{?}
  & 8
  & \tableentry{0.59}{{0.7}} 
  & \tableentry{1.88}{2.8} 
  & \tableentry{{0.50}}{2.9} 
  & \tableentry{1.90}{3.3} 
  & \tableentry{6.67}{{2.8}} 
  & \tableentry{1.22}{2.4} 
  & \tableentry{2.31}{2.5} 
  \\
  \midrule
  \midrule
  \multirow{10}{*}{\STAB{\rotatebox[origin=c]{90}{Image based APR}}}
  & \mbox{PoseNet~\cite{kendall2015posenet}}*
  & 1
  & \tableentry{1.92}{5.4} 
  & \tableentry{2.31}{5.4} 
  & \tableentry{1.46}{8.1} 
  & \tableentry{2.66}{8.5} 
  & - 
  & \tableentry{2.09}{6.8} 
  & - 
  \\
  & \mbox{Bayesian PoseNet~\cite{posenet16}}*
  & 1
  & \tableentry{1.74}{4.1} 
  & \tableentry{2.57}{5.1} 
  & \tableentry{1.25}{7.5} 
  & \tableentry{2.11}{8.4} 
  & - 
  & \tableentry{1.96}{6.0} 
  & - 
  \\
  & \mbox{PN learned weights~\cite{kendall2017geometric}}*
  & 1
  & \tableentry{0.99}{1.1} 
  & \tableentry{2.17}{2.9} 
  & \tableentry{1.05}{4.0} 
  & \tableentry{1.49}{3.4} 
  & \tableentry{7.00}{3.7} 
  & \tableentry{1.43}{2.9} 
  & \tableentry{2.54}{3.0} 
  \\
  & \mbox{Geometric PoseNet~\cite{kendall2017geometric}}*
  & 1
  & \tableentry{0.88}{1.0} 
  & \tableentry{3.20}{3.3} 
  & \tableentry{0.88}{3.8} 
  & \tableentry{1.57}{3.3} 
  & \tableentry{6.83}{3.5} 
  & \tableentry{1.63}{2.7} 
  & \tableentry{2.67}{3.0} 
  \\
  & \mbox{SVS-Pose~\cite{svs_pose}}*
  & 1
  & \tableentry{1.06}{2.8} 
  & \tableentry{1.50}{4.0} 
  & \tableentry{0.63}{5.7} 
  & \tableentry{2.11}{8.1} 
  & - 
  & \tableentry{1.33}{5.2} 
  & - 
  \\
  & \mbox{LSTM-Pose~\cite{poselstm}}*
  & 1
  & \tableentry{0.99}{3.7} 
  & \tableentry{1.51}{4.3} 
  & \tableentry{1.18}{7.4} 
  & \tableentry{1.52}{6.7} 
  & - 
  & \tableentry{1.30}{5.5} 
  & - 
  \\
  & \mbox{GPoseNet~\cite{cai2019hybrid}}*
  & 1
  & \tableentry{1.61}{2.3} 
  & \tableentry{2.62}{3.9} 
  & \tableentry{1.14}{5.7} 
  & \tableentry{2.93}{6.5} 
  & - 
  & \tableentry{2.08}{4.6} 
  & - 
  \\
  & \mbox{MLFBPPose \cite{wang2019discriminative}*}
  & 1
  & \tableentry{0.76}{1.7} 
  & \tableentry{1.99}{2.9} 
  & \tableentry{0.75}{5.1} 
  & \tableentry{{1.29}}{5.0} 
  & - 
  & \tableentry{1.20}{3.7} 
  & - 
  \\
  & \mbox{ADPoseNet \cite{adposenet}*}
  & 1
  & \tableentry{1.30}{1.7} 
  & - 
  & \tableentry{1.22}{6.7} 
  & \tableentry{2.28}{4.8} 
  & - 
  & \tableentry{1.60}{4.2} 
  & - 
  \\
  & \mbox{AnchorPoint~\cite{saha2018improved}*}\textsubscript{?}
  & 1
  & \tableentry{\underline{0.57}}{\textbf{0.9}} 
  & \tableentry{{1.21}}{\underline{2.6}} 
  & \tableentry{\underline{0.52}}{\textbf{2.3}} 
  & \tableentry{\textbf{1.04}}{\textbf{2.7}} 
  & \tableentry{\underline{4.64}}{\underline{3.4}} 
  & \tableentry{\textbf{0.84}}{\textbf{2.1}} 
  & \tableentry{\underline{1.60}}{\underline{2.4}} 
  \\
  & \mbox{AnchorPoint~\cite{saha2018improved}}* publ. code \cite{anchorPointGithub}
  & 1
  & 1.02\textsubscript{2D}, ~-~
  & 0.82\textsubscript{2D}, ~-~ 
  & 0.94\textsubscript{2D}, ~-~ 
  & 1.02\textsubscript{2D}, ~-~ 
  & - 
  & 0.95\textsubscript{2D}, ~-~  
  & - 
  \\
  \midrule
  \multirow{2}{*}{\STAB{\rotatebox[origin=c]{90}{IR}}}
  & \mbox{DenseVLAD~\cite{denseVLAD}}
  & 1
  & \tableentry{2.80}{5.7} 
  & \tableentry{4.01}{7.1} 
  & \tableentry{1.11}{7.6} 
  & \tableentry{2.31}{8.0} 
  & - 
  & \tableentry{2.56}{7.1} 
  & - 
  \\
  & \mbox{DenseVLAD+Inter~\cite{sattler2019understanding}}
  & 1
  & \tableentry{1.48}{4.5} 
  & \tableentry{2.68}{4.6} 
  & \tableentry{0.90}{4.3} 
  & \tableentry{1.62}{6.1} 
  & - 
  & \tableentry{1.67}{4.9} 
  & - 
  \\
   \midrule
  \multirow{2}{*}{\STAB{\rotatebox[origin=c]{90}{RPR}}}
  & \mbox{EssNet~\cite{essnet}}
  & 1
  & \tableentry{0.76}{1.9} 
  & \tableentry{1.39}{2.8} 
  & \tableentry{0.84}{4.3} 
  & \tableentry{1.32}{4.7} 
  & - 
  & \tableentry{1.08}{3.4} 
  & - 
  \\
  & \mbox{NC-EssNet~\cite{essnet}}
  & 1
  & \tableentry{0.61}{1.6} 
  & \tableentry{\textbf{0.95}}{2.7} 
  & \tableentry{0.7\ }{3.4} 
  & \tableentry{\underline{1.12}}{3.6} 
  & - 
  & \tableentry{\underline{0.85}}{2.8} 
  & - 
  \\
  \midrule
  & Ours 
  & 1
  & \tableentry{\textbf{0.48}}{\underline{1.0}} 
  & \tableentry{\underline{1.14}}{\textbf{2.5}} 
  & \tableentry{\textbf{0.48}}{\underline{2.5}} 
  & \tableentry{1.52}{\underline{3.2}} 
  & \tableentry{\textbf{3.2}}{\textbf{2.2}} 
  & \tableentry{0.91}{\underline{2.3}}  
  & \tableentry{\textbf{1.37}}{\textbf{2.3}} 
  \\

\bottomrule
\end{tabular}
\end{table*}

\else
\input{044_table_cambridge}
\fi

\ifmyarxiv
\subsubsection{Generalizability}\label{sec:Generalization}
We experimented to establish how our method generalizes to unseen scenes, as this is a significant virtue of relative pose estimation methods.
Most APR and Sequence-based methods in \Cref{t:comparison,t:cambridge} have no generalization capability to unseen scenes, unlike \mbox{NN-Net}, RelocNet, \mbox{(NC-)EssNet}, IR methods, and ours.
In this experiment, we leave one scene out from the training set and train the model with the remaining six scenes of the 7-Scenes dataset.
Later we evaluate the performance on the $7^{th}$ scene's test set.
In \Cref{t:sixone}, we compare against 
\mbox{NN-Net}, \mbox{EssNet} and \mbox{NC-EssNet}.
We contacted the authors of NN-Net in order to be able to train the remaining four models they didn't publish scores for. The authors were very helpful and provided dataset split files. With their assistance we updated and ran their published code in an attempt to reproduce their published numbers and to produce scores for the unpublished scenes \emph{Fire}, \emph{Office}, \emph{Pumpkin} and \emph{Stairs}.
Also, it was easy to modify the code of (NC-)EssNet to exclude the training set of the scene in the column title.
We verified the correctness of our modifications by reproducing two entries in \Cref{t:comparison} (``EssNet reprod.'' and ``NC-EssNet reprod.'').\\ 
(NC-)EssNet generalizes surprisingly poorly in this evaluation scenario and is significantly outperformed by our method. 
Although \mbox{NN-Net}'s published results are impressive, its performance on translation for some cases, \eg \emph{Fire}, \emph{Office}, is bad.
Overall, in this generalization experiment, our method either outperforms \mbox{NN-Net} or performs on par while being simpler and faster to train than RPR methods that rely on metric learning strategies.
For instance, NN-Net training with 7-Scenes takes $\approx 3$ days on two V100 GPUs \vs our comparable performance after $\approx 1$ hour on a single V100. Specifically, training our model with the entire 7-Scenes dataset takes $< 20$ hours.
NN-Net training diverged for two of the scenes (\emph{Kitchen}, \emph{Heads}). For those scenes we report the performance of the last checkpoint before divergence.\\
Lastly, note that our method performs better than PoseNet~\cite{kendall2015posenet} on translation, even though it has not seen the scene before.
\else
\subsubsection{Generalizability}\label{sec:Generalization}
We experimented to establish how our method generalizes to unseen scenes, as this is a significant virtue of relative pose estimation methods.
Most APR and Sequence-based methods in \Cref{t:comparison,t:cambridge} have no generalization capability to unseen scenes, unlike \mbox{NN-Net}, RelocNet, \mbox{NC-EssNet}, IR methods, and ours.
In this experiment, we leave one scene out from the training set and train the model with the remaining six scenes of the 7-Scenes dataset.
Later we evaluate the performance on the $7^{th}$ scene's test set.
In \Cref{t:sixone}, we compare against 
\mbox{NN-Net}.
The proposed method shows similar performance to \mbox{NN-Net} on the translation metric, and some improvement when measuring accuracy of orientation estimates. Further experiments are shown in the Supplementary Material. Note that the proposed method and \mbox{NN-Net} perform better than PoseNet~\cite{kendall2015posenet} on translation, even though they haven't seen the scene before. 
Our model is more accurate and simpler to train than RPR methods that rely on metric learning strategies.
For instance, NN-Net training with 7-Scenes takes $\approx 3$ days on two V100 GPUs \vs our comparable performance after $\approx 1$ hour on a single V100. Specifically, training our model with the entire 7-Scenes dataset takes $< 20$ hours.
\fi

\subsection{Ablation Study}\label{sec:ablation}
\begin{table*}[t]
\centering
\caption{{\bf Ablation study.} We evaluate the importance of the individual ingredients of our method. \emph{Baseline-1} (without GNN) consists of FCs transforming the concatenated node features of an image pair. \emph{Baseline-2} (without image retrieval) builds a graph at test time from random images from the training set instead of the most relevant images in our method. \emph{Baseline-3} (without image retrieval, sequence) simulates GL-Net \cite{glnet} by building a graph from consecutive test frames.}\label{t:ablation}
\small
\setlength\tabcolsep{2pt} 
\begin{tabular}{lccccccccc}
\midrule
 & \begin{tabular}{@{}c@{}} \# test frames \end{tabular} & Chess & Fire & Heads & Office & Pumpkin & Kitchen & Stairs & Avg. \\
\toprule
\mbox{Baseline-1: w/o GNN}
  & 1
  & \tableentry{0.30}{14.2} 
  & \tableentry{0.43}{18.2} 
  & \tableentry{0.19}{18.1} 
  & \tableentry{0.41}{13.9} 
  & \tableentry{0.47}{15.0} 
  & \tableentry{0.41}{15.1} 
  & \tableentry{0.32}{16.2} 
  & \tableentry{0.36}{15.8} 
  \\
  \midrule
  \mbox{Baseline-2: w/o IR, rand.}
  & 1
  & \tableentry{0.14}{4.3} 
  & \tableentry{0.27}{9.7} 
  & \tableentry{0.15}{9.1} 
  & \tableentry{0.21}{5.7} 
  & \tableentry{0.23}{5.0} 
  & \tableentry{0.27}{5.2} 
  & \tableentry{0.28}{7.9} 
  & \tableentry{0.22}{6.7} 
  \\
   Baseline-3: w/o IR, seq \textapprox \cite{glnet} 
  & 8
  & \tableentry{0.20}{6.8} 
  & \tableentry{0.35}{12.3} 
  & \tableentry{0.18}{14.2} 
  & \tableentry{0.26}{7.7} 
  & \tableentry{0.31}{6.9} 
  & \tableentry{0.36}{7.9} 
  & \tableentry{0.43}{13.1} 
  & \tableentry{0.30}{9.8} 
  \\
  \midrule
    Ours
  & 1
  & \tableentry{\textbf{0.08}}{\textbf{2.7}} 
  & \tableentry{\textbf{0.21}}{\textbf{7.5}} 
  & \tableentry{\textbf{0.13}}{\textbf{8.7}} 
  & \tableentry{\textbf{0.15}}{\textbf{4.1}} 
  & \tableentry{\textbf{0.15}}{\textbf{3.5}} 
  & \tableentry{\textbf{0.19}}{\textbf{3.7}} 
  & \tableentry{\textbf{0.22}}{\textbf{6.5}} 
  & \tableentry{\textbf{0.16}}{\textbf{5.2}} 
  \\

\bottomrule
\end{tabular}
\end{table*}

To show the effectiveness of GNNs for relative pose learning, we compare our approach to multiple baseline models. For the first baseline, denoted \emph{Baseline-1}, we remove the GNN and instead add fully connected layers (FCs) on top of the visual feature extractor, such that both models have similar numbers of parameters.
This baseline is essentially the same as PoseNet, except that it regresses relative rather than absolute poses. It takes two feature vectors, concatenates them, and regresses the relative poses between the corresponding image pairs from their joint embedding after the FC layers. We compare the results for 7-Scenes in \Cref{t:ablation}. The comparison clearly shows that the GNN is essential to estimate the relative pose in our simple, ``direct'' setting, without metric learning tricks \cite{relocnet}.

We conducted another experiment to illustrate the importance of image retrieval in our method.
We removed the image retrieval module to obtain two baselines, \emph{Baseline-2} and \emph{Baseline-3}.
In the first case, instead of constructing the graph via image retrieval, we construct it by random sampling.
For each scene, we randomly sample 8 images and link them into a fully connected graph. 
In the second case, we construct the graph from sequential frames, as in \mbox{GL-Net~\cite{glnet}}. 
We train this model with both absolute and relative pose losses, and at test time we estimate the absolute poses of the test images directly from the node features.
The main difference compared to \cite{glnet} is that \cite{glnet} includes a multi-scale approach and incorporates 2D convolutional layers inside the GNN.
\Cref{t:ablation} compares the results on 7-Scenes.
Results somewhat surprisingly show that using a GNN is still beneficial, even without an image retrieval module, however it is clear that it is an important complement to GNNs and they work best together. 
One can observe that the performance difference between Baseline-2 and the proposed method is smallest for the \emph{Heads} scene. This is because the \emph{Heads} scene is small and the camera only looks at a limited region of the scene. Therefore, the random sampling approach also provides reasonably related and diverse set of images for the graph construction.
In contrast, the performance gap is larger for large scenes \eg \emph{Office}, thereby supporting the idea that 
image retrieval in our method gets more useful as the scene size increases.

\subsection{Implementation Details}

We implemented our method using the PyTorch Geometric library for GNNs. For all experiments, we use Adam~\cite{adam} as an optimizer, with batch size 8, and run the training for 50 epochs. The learning rate is initially set to $5\times10^{-5}$ and divided by 10 every 20 epochs. The model is regularized with weight decay of $5\times10^{-4}$ and early stopping. The edge dropout factor is set to 0.5, images are resized to a height of 256 pixels, maintaining the aspect ratio. 
\ifmyarxiv
Source code and our trained models will be posted online upon acceptance.
\else
Source code is submitted in the supplementary materials, and will be posted online with our trained models.
\fi

\section{Conclusion \& Future Work}

In this paper, we have shown that graph neural networks combined with an image retrieval module and supervised only with a relative pose loss outperform APR and sequence-based methods in camera re-localization. 
Training using only relative pose supervision enables training on multiple datasets, yielding robustness towards overfitting, and enabling generalization to unseen scenes, as shown in \Cref{t:comparison}. In ablation experiments, we have shown that message passing is indeed necessary.
Further, training GNNs is much simpler and faster than earlier RPR methods (\eg RelocNet), whilst achieving better performance.

GNNs are good at encoding dynamic and sparse neighborhoods, hence constructing graphs with different neighborhood sizes is on our future roadmap. Especially appealing, we plan to explore learning sparse graph representations of view-points with relative pose supervision.

{\small
\bibliographystyle{ieee_fullname}
\bibliography{RelPoseGraphs}
}

\ifmyarxiv
\clearpage
\begin{center}
\textbf{\huge Supplementary Material}
\setcounter{section}{0}
\end{center}
\ifmyarxiv
\else
\section{Updates to Tables 1 and 3 in the paper}
The authors of \cite{essnet} were kind enough to provide us with detailed numbers for their method, where we only had the averages before. See ``EssNet'' and ``NC-EssNet'' in \Cref{t:supp:comparison,t:supp:cambridge}.
We also validated our use of their published code by attempting to reproduce their published numbers, when training on the training set of 7-Scenes. We were happy to find, that we can indeed get close to the numbers they reported. We therefore conclude, that our use of their published code in \Cref{t:supp:comparisonGeneralization} is correct.

\begin{table*}[!ht]
\centering
\caption[Comparison to state-of-the-art methods: 7-Scenes.]{{\bf Comparison to state-of-the-art methods: 7-Scenes.}
Median translation and rotation error. Methods marked with a * have scene-specific representation and do not generalize to unseen scenes. 
We've marked methods with difficult reproducibility with ''?''.
Per-scene results provided by the authors of \mbox{NC-EssNet~\cite{essnet}} added. We also successfully validated those scores using their published code (\cf ``EssNet reprod. Stairs'' and ``NC-EssNet reprod. Chess'').
Shortened version for readability, entries unchanged compared to paper in grey.
}
\label{t:supp:comparison}
\small
\setlength\tabcolsep{2pt} 
\begin{tabular}{clccccccccc}
\midrule
 & & $\!\!\!\!\!\!\!$\# test frames & Chess & Fire & Heads & Office & Pumpkin & Kitchen & Stairs & Avg. \\
\toprule
  & \textcolor{inpaper}{\mbox{DSAC\textsuperscript{*}~\cite{brachmann2020dsacstar}}*}
  & \textcolor{inpaper}{1}
  & \textcolor{inpaper}{\tableentry{0.02}{1.1}} 
  & \textcolor{inpaper}{\tableentry{0.02}{1.2}} 
  & \textcolor{inpaper}{\tableentry{0.01}{1.8}}
  & \textcolor{inpaper}{\tableentry{0.03}{1.2}}
  & \textcolor{inpaper}{\tableentry{0.04}{1.4}}
  & \textcolor{inpaper}{\tableentry{0.03}{1.7}}
  & \textcolor{inpaper}{\tableentry{0.04}{1.4}}
  & \textcolor{inpaper}{\tableentry{0.03}{1.4}}
  \\
  \noalign{\hrule height 1pt}
  & \textcolor{inpaper}{GL-Net~\cite{glnet}*\textsubscript{?}}
  & \textcolor{inpaper}{8}
  & \textcolor{inpaper}{\tableentry{{0.08}}{{2.8}}} 
  & \textcolor{inpaper}{\tableentry{0.26}{{8.9}}} 
  & \textcolor{inpaper}{\tableentry{0.17}{11.4}} 
  & \textcolor{inpaper}{\tableentry{0.18}{{5.1}}} 
  & \textcolor{inpaper}{\tableentry{{0.15}}{{2.8}}} 
  & \textcolor{inpaper}{\tableentry{0.25}{{4.5}}} 
  & \textcolor{inpaper}{\tableentry{0.23}{8.8}} 
  & \textcolor{inpaper}{\tableentry{0.19}{{6.3}}} 
  \\
  \midrule
  & \textcolor{inpaper}{\mbox{AnchorPoint~\cite{saha2018improved}*}\textsubscript{?}}
  & \textcolor{inpaper}{1}
  & \textcolor{inpaper}{\tableentry{\textbf{0.06}}{\underline{3.9}}} 
  & \textcolor{inpaper}{\tableentry{\textbf{0.16}}{11.1}} 
  & \textcolor{inpaper}{\tableentry{\textbf{0.09}}{11.2}} 
  & \textcolor{inpaper}{\tableentry{\textbf{0.11}}{5.4}} 
  & \textcolor{inpaper}{\tableentry{\textbf{0.14}}{\underline{3.6}}} 
  & \textcolor{inpaper}{\tableentry{\textbf{0.13}}{5.3}} 
  & \textcolor{inpaper}{\tableentry{\textbf{0.21}}{11.9}} 
  & \textcolor{inpaper}{\tableentry{\underline{0.13}}{7.5}} 
  \\ 
  \midrule 
  \multirow{7}{*}{\STAB{\rotatebox[origin=c]{90}{RPR}}}
  & \textcolor{inpaper}{\mbox{NN-Net~\cite{nn-net}}}
  & \textcolor{inpaper}{1}
  & \textcolor{inpaper}{\tableentry{0.13}{6.5}} 
  & \textcolor{inpaper}{\tableentry{0.26}{12.7}} 
  & \textcolor{inpaper}{\tableentry{0.14}{12.3}} 
  & \textcolor{inpaper}{\tableentry{0.21}{7.4}} 
  & \textcolor{inpaper}{\tableentry{0.24}{6.4}} 
  & \textcolor{inpaper}{\tableentry{0.24}{8.0}} 
  & \textcolor{inpaper}{\tableentry{0.27}{11.8}} 
  & \textcolor{inpaper}{\tableentry{0.21}{9.3}} 
  \\
  & \textcolor{inpaper}{\mbox{RelocNet~\cite{relocnet}}}
  & \textcolor{inpaper}{1}
  & \textcolor{inpaper}{\tableentry{0.12}{4.1}} 
  & \textcolor{inpaper}{\tableentry{0.26}{10.4}} 
  & \textcolor{inpaper}{\tableentry{0.14}{\underline{10.5}}} 
  & \textcolor{inpaper}{\tableentry{0.18}{\underline{5.3}}} 
  & \textcolor{inpaper}{\tableentry{0.26}{4.2}} 
  & \textcolor{inpaper}{\tableentry{0.23}{\underline{5.1}}} 
  & \textcolor{inpaper}{\tableentry{0.28}{\underline{7.5}}} 
  & \textcolor{inpaper}{\tableentry{0.21}{6.7}} 
  \\
  & \mbox{EssNet~\cite{essnet}}
  & 1
  & \tableentry{0.13}{5.1} 
  & \tableentry{0.27}{10.1} 
  & \tableentry{0.15}{9.9} 
  & \tableentry{0.21}{6.9} 
  & \tableentry{0.22}{6.1} 
  & \tableentry{0.23}{6.9} 
  & \tableentry{0.32}{11.2} 
  & \textcolor{inpaper}{\tableentry{0.22}{8.0}} 
  \\
      & \mbox{EssNet~\cite{essnet} reprod.}
  & 1
  & - 
  & - 
  & - 
  & - 
  & - 
  & - 
  & \tableentry{0.32}{9.8} 
  & - 
  \\
    & \mbox{NC-EssNet~\cite{essnet}}
  & 1
  & \tableentry{0.12}{5.6} 
  & \tableentry{0.26}{9.6} 
  & \tableentry{0.14}{10.7} 
  & \tableentry{0.2\ }{6.7} 
  & \tableentry{0.22}{5.7} 
  & \tableentry{0.22}{6.3} 
  & \tableentry{0.31}{7.9} 
  & \textcolor{inpaper}{\tableentry{0.21}{7.5}} 
  \\
     & \mbox{NC-EssNet~\cite{essnet} reprod.}
  & 1
  & \tableentry{0.13}{5.5} 
  & - 
  & - 
  & - 
  & - 
  & - 
  & - 
  & - 
  \\
  & \textcolor{inpaper}{CamNet \cite{camnet}*\textsubscript{?}}
  & \textcolor{inpaper}{1}
  & \textcolor{inpaper}{-} 
  & \textcolor{inpaper}{-} 
  & \textcolor{inpaper}{-} 
  & \textcolor{inpaper}{-} 
  & \textcolor{inpaper}{-} 
  & \textcolor{inpaper}{-} 
  & \textcolor{inpaper}{-} 
  & \textcolor{inpaper}{\tableentry{\textbf{0.05}}{\textbf{1.8}}} 
  \\ 
\midrule
  & \textcolor{inpaper}{Ours} 
  & \textcolor{inpaper}{1}
  & \textcolor{inpaper}{\tableentry{\underline{0.08}}{\textbf{2.7}}} 
  & \textcolor{inpaper}{\tableentry{\underline{0.21}}{\textbf{7.5}}} 
  & \textcolor{inpaper}{\tableentry{\underline{0.13}}{\textbf{8.7}}} 
  & \textcolor{inpaper}{\tableentry{\underline{0.15}}{\textbf{4.1}}} 
  & \textcolor{inpaper}{\tableentry{\underline{0.15}}{\textbf{3.5}}} 
  & \textcolor{inpaper}{\tableentry{\underline{0.19}}{\textbf{3.7}}} 
  & \textcolor{inpaper}{\tableentry{\underline{0.22}}{\textbf{6.5}}} 
  & \textcolor{inpaper}{\tableentry{0.16}{\underline{5.2}}} 
  \\
\bottomrule
\end{tabular}
\end{table*}

\fi

\ifmyarxiv
\else
\section{Comparison of capability to generalize}

In \Cref{t:sixone} we compare our performance to the state-of-the-art visual relative pose estimation methods \mbox{NC-EssNet~\cite{essnet}} and \mbox{NN-Net~\cite{nn-net}}. 
It was easy to modify the excellent quality code of NC-EssNet to exclude the training set of the scene in the column title.
We verified the correctness of our modifications by reproducing two entries in \Cref{t:comparison} (``EssNet reprod.'' and ``NC-EssNet reprod.''). 
NC-EssNet generalizes surprisingly badly in this evaluation scenario of not training on the training set of the test set. \\
We contacted the authors of NN-Net in order to be able to train the remaining four models they didn't publish scores for. 
The authors were very helpful and provided dataset split files. With their assistance we updated and ran their published code in an attempt to reproduce their published numbers and to produce scores for the unpublished scenes Fire, Office, Pumpkin and Stairs.
Overall, their published results are impressive, but need more than two days of training on two V100 GPUs. 
Our method can be trained on a single V100 GPU in less than a day, and gives more reliable results.
The NN-Net training processes diverged for two of the scenes (Kitchen, Heads), so for those scenes we report the performance of the last checkpoint before the explosion. \\
The differences in the smoothness and stability of the relative pose estimates in this generalization setting are even more eye-catching in the supplementary video. We use the estimated relative poses and the depths (\emph{only} for rendering) in the datasets to reproject the query images into the estimated camera locations. Would the estimated poses be perfect, we should only see differences between the input RGB images and the reprojected depths due to missing depth information.

\begin{table*}[!h]
\centering
\caption{{\bf Generalization to an unseen environment.} All methods were trained on the training set of six scenes, and evaluated on the test set of the 7th scene. We report median translation and rotation errors. The authors of \protect{\cite{nn-net}} provided us with their dataset files describing the dataset splits for the unpublished cases and we used their published training and evaluation code to reproduce their results. The training on the Heads and Kitchen scenes did not converge$^{\dagger}$ in their case, we used the best models before divergence. Overall, our method produces more reliable results with a faster training procedure and only \emph{one} GPU. 
We modified the published code for \cite{essnet} to produce their results. They perform surprisingly badly in this setting, perhaps the number of reference images (one, compared to NN-Net's five and our seven) is their main limitation.
}\label{t:supp:comparisonGeneralization}
\small
\setlength\tabcolsep{1pt} 
\begin{tabular}{lcccccccc|ccc}
\midrule
 & Chess & Fire & Heads & Office & Pumpkin & Kitchen & Stairs &&
 & \begin{tabular}{@{}c@{}} Avg. (3 scenes) \end{tabular}  
 & Avg.
 \\
\toprule
\textcolor{inpaper}{\mbox{NN-Net~\cite{nn-net}} published}
  & \textcolor{inpaper}{\tableentry{\textbf{0.27}}{13.1}} 
  & \textcolor{inpaper}{-} 
  & \textcolor{inpaper}{\tableentry{0.23}{15.0}} 
  & \textcolor{inpaper}{-}
  & \textcolor{inpaper}{-}
  & \textcolor{inpaper}{\tableentry{\textbf{0.36}}{12.6}} 
  & \textcolor{inpaper}{-}
  &&& \textcolor{inpaper}{\tableentry{{0.29}}{13.6}} 
  & \textcolor{inpaper}{-} 
\\
\midrule
\mbox{EssNet~\cite{essnet}} reprod.
& \tableentry{0.73}{37.6} 
& \tableentry{0.89}{67.6} 
& \tableentry{0.62}{28.5} 
& \tableentry{0.84}{36.3} 
& \tableentry{1.06}{33.3} 
& \tableentry{0.91}{36.1} 
& \tableentry{1.19}{42.1} 
&&
& \tableentry{0.75}{34.1} 
& \tableentry{0.89}{40.2} 
\\
\mbox{NC-EssNet~\cite{essnet}} reprod.
& \tableentry{0.62}{24.2} 
& \tableentry{0.75}{23.7} 
& \tableentry{0.44}{25.6} 
& \tableentry{0.88}{28.0} 
& \tableentry{1.02}{24.5} 
& \tableentry{0.77}{20.8} 
& \tableentry{1.25}{36.5} 
&&
& \tableentry{0.61}{23.5} 
& \tableentry{0.82}{26.2} 
\\
\textcolor{inpaper}{\mbox{NN-Net~\cite{nn-net}} reprod.}
  & \textcolor{inpaper}{\tableentry{0.33}{\textbf{11.5}}} 
  & \textcolor{inpaper}{\tableentry{11.68}{\textbf{13.8}}} 
  & \textcolor{inpaper}{0.30$^{\dagger}$, 15.5\textdegree$^{\dagger}$} 
  & \textcolor{inpaper}{\tableentry{1.34}{\textbf{10.9}}} 
  & \textcolor{inpaper}{\tableentry{\textbf{0.41}}{12.8}} 
  & \textcolor{inpaper}{1.68$^{\dagger}$, 12.9\textdegree$^{\dagger}$} 
  & \textcolor{inpaper}{\tableentry{0.44}{\textbf{13.6}}} 
  &&
  & \textcolor{inpaper}{\tableentry{0.77}{\textbf{13.2}}} 
  & \textcolor{inpaper}{\tableentry{2.31}{\textbf{13.0}}} 
\\
  \textcolor{inpaper}{Ours} 
  & \textcolor{inpaper}{\tableentry{{0.29}}{12.8}}

  & \textcolor{inpaper}{\tableentry{\textbf{0.45}}{15.7}}

  & \textcolor{inpaper}{\tableentry{\textbf{0.19}}{\textbf{14.7}}}
  & \textcolor{inpaper}{\tableentry{\textbf{0.42}}{12.5}}

  & \textcolor{inpaper}{\tableentry{0.44}{\textbf{11.7}}}
  & \textcolor{inpaper}{\tableentry{{0.42}}{\textbf{12.4}}}
  & \textcolor{inpaper}{\tableentry{\textbf{0.35}}{15.5}}
  &&
  & \textcolor{inpaper}{\tableentry{{0.30}}{13.3}} 
  & \textcolor{inpaper}{\tableentry{\textbf{0.37}}{13.6}} 
  \\
\bottomrule
\end{tabular}
\end{table*}

\begin{table*}[!ht]
\centering
\caption{{\bf Comparison to state-of-the-art methods: Cambridge Landmarks.} 
Median translation and rotation error. Methods marked with a * have scene-specific representation and do not generalize to unseen scenes. ''?'' indicates methods with difficult reproducibility.
The authors of \cite{essnet} provided us with their detailed results.
Shortened version for readability, entries unchanged compared to paper in grey.
}\label{t:supp:cambridge}
\small
\setlength\tabcolsep{3pt} 
\begin{tabular}{llcccccccc}
\midrule
& & \# test frames & College & Hospital & Shop & Church & Court 
 & \begin{tabular}{@{}c@{}}  Avg. (4) \end{tabular} & \begin{tabular}{@{}c@{}}  Avg. (5) \end{tabular} \\
\toprule
  & \textcolor{inpaper}{\mbox{DSAC\textsuperscript{*}~\cite{brachmann2020dsacstar}}*}
  & \textcolor{inpaper}{1}
  & \textcolor{inpaper}{\tableentry{0.18}{0.3}} 
  & \textcolor{inpaper}{\tableentry{0.21}{0.4}} 
  & \textcolor{inpaper}{\tableentry{0.05}{0.3}} 
  & \textcolor{inpaper}{\tableentry{0.15}{0.5}} 
  & \textcolor{inpaper}{\tableentry{0.34}{0.2}} 
  & \textcolor{inpaper}{\tableentry{0.15}{0.4}} 
  & \textcolor{inpaper}{\tableentry{0.19}{0.3}} 
  \\
  \noalign{\hrule height 1pt}
  & \textcolor{inpaper}{\mbox{GL-Net~\cite{glnet}*}\textsubscript{?}}
  & \textcolor{inpaper}{8}
  & \textcolor{inpaper}{\tableentry{0.59}{{0.7}}} 
  & \textcolor{inpaper}{\tableentry{1.88}{2.8}} 
  & \textcolor{inpaper}{\tableentry{{0.50}}{2.9}} 
  & \textcolor{inpaper}{\tableentry{1.90}{3.3}} 
  & \textcolor{inpaper}{\tableentry{6.67}{{2.8}}} 
  & \textcolor{inpaper}{\tableentry{1.22}{2.4}} 
  & \textcolor{inpaper}{\tableentry{2.31}{2.5}} 
  \\
  \midrule
  & \textcolor{inpaper}{\mbox{AnchorPoint~\cite{saha2018improved}*}\textsubscript{?}}
  & \textcolor{inpaper}{1}
  & \textcolor{inpaper}{\tableentry{\underline{0.57}}{\textbf{0.9}}} 
  & \textcolor{inpaper}{\tableentry{{1.21}}{\underline{2.6}}} 
  & \textcolor{inpaper}{\tableentry{\underline{0.52}}{\textbf{2.3}}} 
  & \textcolor{inpaper}{\tableentry{\textbf{1.04}}{\textbf{2.7}}} 
  & \textcolor{inpaper}{\tableentry{\underline{4.64}}{\underline{3.4}}} 
  & \textcolor{inpaper}{\tableentry{\textbf{0.84}}{\textbf{2.1}}} 
  & \textcolor{inpaper}{\tableentry{\underline{1.60}}{\underline{2.4}}} 
  \\
  & \textcolor{inpaper}{\mbox{AnchorPoint~\cite{saha2018improved}}* reprod. \cite{anchorPointGithub}}
  & \textcolor{inpaper}{1}
  & \textcolor{inpaper}{1.02\textsubscript{2D}, ~-~}
  & \textcolor{inpaper}{0.82\textsubscript{2D}, ~-~} 
  & \textcolor{inpaper}{0.94\textsubscript{2D}, ~-~} 
  & \textcolor{inpaper}{1.02\textsubscript{2D}, ~-~} 
  & \textcolor{inpaper}{-} 
  & \textcolor{inpaper}{0.95\textsubscript{2D}, ~-~}  
  & \textcolor{inpaper}{-} 
  \\
   \midrule
  \multirow{2}{*}{\STAB{\rotatebox[origin=c]{90}{RPR}}}
  & \mbox{EssNet~\cite{essnet}}
  & 1
  & \tableentry{0.76}{1.9} 
  & \tableentry{1.39}{2.8} 
  & \tableentry{0.84}{4.3} 
  & \tableentry{1.32}{4.7} 
  & - 
  & \tableentry{1.08}{3.4} 
  & - 
  \\
  & \mbox{NC-EssNet~\cite{essnet}}
  & 1
  & \tableentry{0.61}{1.6} 
  & \tableentry{\textbf{0.95}}{2.7} 
  & \tableentry{0.7\ }{3.4} 
  & \tableentry{\underline{1.12}}{3.6} 
  & - 
  & \tableentry{\underline{0.85}}{2.8} 
  & - 
  \\
  \midrule
  & \textcolor{inpaper}{Ours}
  & \textcolor{inpaper}{1}
  & \textcolor{inpaper}{\tableentry{\textbf{0.48}}{\underline{1.0}}} 
  & \textcolor{inpaper}{\tableentry{\underline{1.14}}{\textbf{2.5}}} 
  & \textcolor{inpaper}{\tableentry{\textbf{0.48}}{\underline{2.5}}} 
  & \textcolor{inpaper}{\tableentry{1.52}{\underline{3.2}}} 
  & \textcolor{inpaper}{\tableentry{\textbf{3.2}}{\textbf{2.2}}} 
  & \textcolor{inpaper}{\tableentry{0.91}{\underline{2.3}}}  
  & \textcolor{inpaper}{\tableentry{\textbf{1.37}}{\textbf{2.3}}} 
  \\
\bottomrule
\end{tabular}
\end{table*}

\fi

\section{Single-scene Training}

In \Cref{t:single} we compare our solution trained only on a single scene to our strongest competitors, including those performing Absolute Pose Regression (APR), which usually score higher than Relative Pose Regression (RPR) models. Although our use of only a relative pose loss allows us to benefit from the richness of data in multiple scenes, even our single-scene model shows competitive performance in the standard benchmarks.

\begin{table*}[!t]
\centering
\caption{{\bf Single-scene training: 7-Scenes.} We compare training our method on a single scene's training set (``Ours single'') to training our method on all scenes' training sets. Note, that even our single scene version finishes among the top-3 methods when compared to our closest competitors.
We marked the best performing result \colorbox{gold}{gold}, the second best \colorbox{silver}{silver}, and the third best performing result \colorbox{bronze}{bronze}.}\label{t:single}
\small
\setlength\tabcolsep{3pt} 
\begin{tabular}{llcccccccccccccccc|ccc}
\midrule
 & 
 & \begin{tabular}{@{}c@{}} \# test \\ frames \end{tabular} 
 & \multicolumn{2}{c}{Chess} 
 & \multicolumn{2}{c}{Fire}
 & \multicolumn{2}{c}{Heads} 
 & \multicolumn{2}{c}{Office} 
 & \multicolumn{2}{c}{Pumpkin} 
 & \multicolumn{2}{c}{Kitchen} 
 & \multicolumn{2}{c}{Stairs} 
 &&\multicolumn{1}{|c}{} & \multicolumn{2}{c}{Avg.} \\
\toprule
\multirow{2}{*}{\STAB{\rotatebox[origin=c]{90}{RPR}}}
& NN-Net~\cite{nn-net} & 1 & 0.13 & 6.5 & 0.26 & 12.7 & \rankthird{0.14} & 12.3 & 0.21 & 7.4 & 0.24 & 6.4 & 0.24 & 8.0 & 0.27 & 11.8 &&\multicolumn{1}{|c}{} & 0.21 & 9.3 \\
& RelocNet~\cite{relocnet} & 1 & 0.12 & 4.1 & 0.26 & 10.4 & \rankthird{0.14} & \ranksecond{10.5} & 0.18 & 5.3 & 0.26 & 4.2 & 0.23 & 5.1 & 0.28 & \ranksecond{7.5} && \multicolumn{1}{|c}{} & 0.21 & 6.7 \\
\midrule
\multirow{2}{*}{\STAB{\rotatebox[origin=c]{90}{APR}}}
& GL-Net~\cite{glnet}* & 8 & \ranksecond{0.08} & \ranksecond{2.8} & 0.26 & \ranksecond{8.9} & 0.17 & 11.4 & 0.18 & \rankthird{5.1} & \ranksecond{0.15} & \rankfirst{2.8} & 0.25 & \rankthird{4.5} & \rankthird{0.23} & 8.8 &&\multicolumn{1}{|c}{} & 0.19 & \ranksecond{6.3} \\
& AnchorPoint~\cite{saha2018improved}* & 1 & \rankfirst{0.06} & 3.9 & \rankfirst{0.16} & 11.1 & \rankfirst{0.09} & \rankthird{11.2} & \rankfirst{0.11} & 5.4 & \rankfirst{0.14} & \rankthird{3.6} & \rankfirst{0.13} & 5.3 & \rankfirst{0.21} & 11.9 &&\multicolumn{1}{|c}{} & \rankfirst{0.13} & 7.5 \\
\midrule
& Ours single & 1 & 0.09 & \ranksecond{2.8} & \rankthird{0.24} & \rankthird{9.2} & 0.15 & 12.6 & \rankthird{0.16} & \ranksecond{4.6} & 0.16 & \rankthird{3.6} & \rankthird{0.21} & \ranksecond{4.3} & 0.25 & \rankthird{8.5} &&\multicolumn{1}{|c}{}& \rankthird{0.18} & \rankthird{6.5} \\
& Ours & 1 & \ranksecond{0.08} & \rankfirst{2.7} & \ranksecond{0.21} & \rankfirst{7.5} & \ranksecond{0.13} & \rankfirst{8.7} & \ranksecond{0.15} & \rankfirst{4.1} & \ranksecond{0.15} & \ranksecond{3.5} & \ranksecond{0.19} & \rankfirst{3.7} & \ranksecond{0.22} & \rankfirst{6.5} &&\multicolumn{1}{|c}{}& \ranksecond{0.16} & \rankfirst{5.2} \\
\bottomrule
\end{tabular}
\end{table*}

\section{Geometric Averaging}
\begin{table*}[!t]
\centering
\caption{{\bf Geometric Averaging (GA): 7-Scenes.} Robust geometric averaging can be used as a post-processing step in the proposed method. We compare the proposed method (Ours) with the proposed method combined with robust geometric averaging (Ours w/ GA). It generally boosts the performance in the rotation metric; however, its contribution to final performance is little compared to the main ingredients of the proposed method (See Section 4.2 in the main text for the ablation study). }\label{t:geomavg}
\small
\setlength\tabcolsep{4pt} 
\begin{tabular}{lccccccccc|cc}
\midrule
 & \begin{tabular}{@{}c@{}} \# test \\ frames \end{tabular} & Chess & Fire & Heads & Office & Pumpkin & Kitchen & Stairs &&& Avg. \\
\toprule
\mbox{Ours}
    & 1
  & \tableentry{\textbf{0.08}}{2.68} 
  & \tableentry{\textbf{0.21}}{7.53} 
  & \tableentry{\textbf{0.13}}{\textbf{8.72}} 
  & \tableentry{\textbf{0.15}}{4.08} 
  & \tableentry{\textbf{0.15}}{3.47} 
  & \tableentry{\textbf{0.19}}{3.67} 
  & \tableentry{\textbf{0.22}}{\textbf{6.51}} 
  &&& \tableentry{\textbf{0.16}}{5.24} 
  \\
  \midrule
    Ours w/ GA 
  & 1
  & \tableentry{\textbf{0.08}}{\textbf{2.63}} 
  & \tableentry{\textbf{0.21}}{\textbf{7.48}} 
  & \tableentry{\textbf{0.13}}{8.75} 
  & \tableentry{\textbf{0.15}}{\textbf{4.00}} 
  & \tableentry{\textbf{0.15}}{\textbf{3.46}} 
  & \tableentry{\textbf{0.19}}{\textbf{3.56}} 
  & \tableentry{\textbf{0.22}}{6.61} 
  &&& \tableentry{\textbf{0.16}}{\textbf{5.21}} 
  \\

\bottomrule
\end{tabular}
\end{table*}

For the absolute pose of the query image, we obtain $N\!-\!1=7$ estimates in our method. It is natural to compute the final estimate by simply taking the average. We conduct an experiment to see how effective robust geometric averaging is. We use Weiszfeld's algorithm for translation and \cite{markley2007averaging} for computing orientations as a post-processing step.
See Table~\ref{t:geomavg} for the comparison. It shows that overall, averaging multiple neighbors slightly improves scores on the rotation metric. We observe only a small performance difference between choosing a random estimate out of $N\!-\!1=7$ estimates or averaging over $N\!-\!1=7$ estimates. This is due to small variance across our model's estimates.
We do not include geometric averaging as a part of our method since it is much less impactful compared to the main components of our approach (see Section 4.2 in the main text for an ablation study). However, sacrificing simplicity, rotation averaging could be used to improve results slightly.

\section{Qualitative Results}

In \Cref{fig:qualitative}, we show some qualitative results over the test set of the \emph{Kitchen} scene from the 7-Scenes dataset. The ranking of estimates was performed according to \Cref{eq:loss}. Our estimates are shown in blue, while the ground truth is shown in green. Note, how the median error examples, shown in the middle row, are still quite close to their ground truth equivalents.

\begin{figure*}[t]
    \centering
    \begin{overpic}[width=0.32\textwidth]{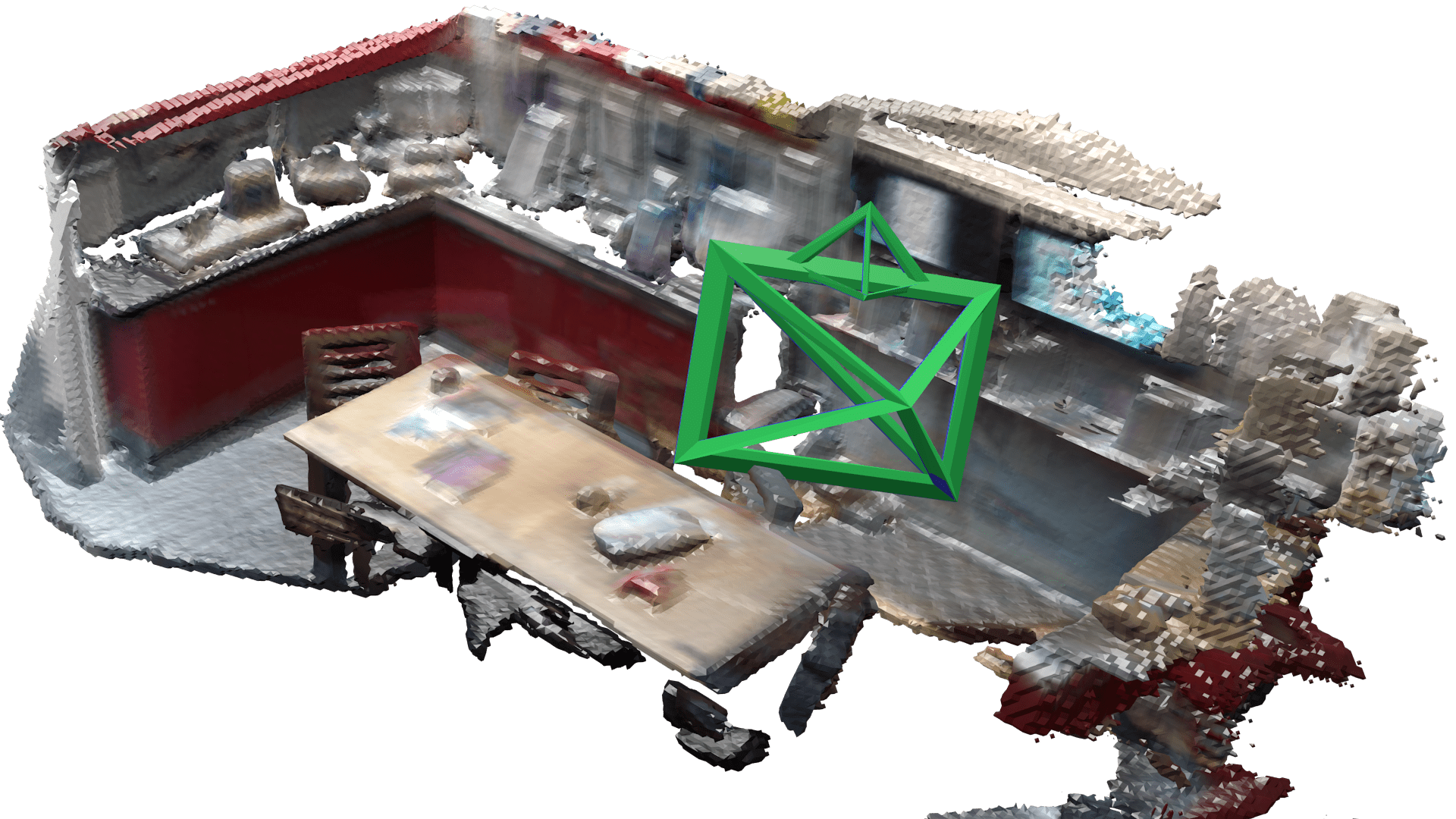}
      \put(0, 54){Best}
      \put(0, 5){\#1}
    \end{overpic}
    \begin{overpic}[width=0.32\textwidth]{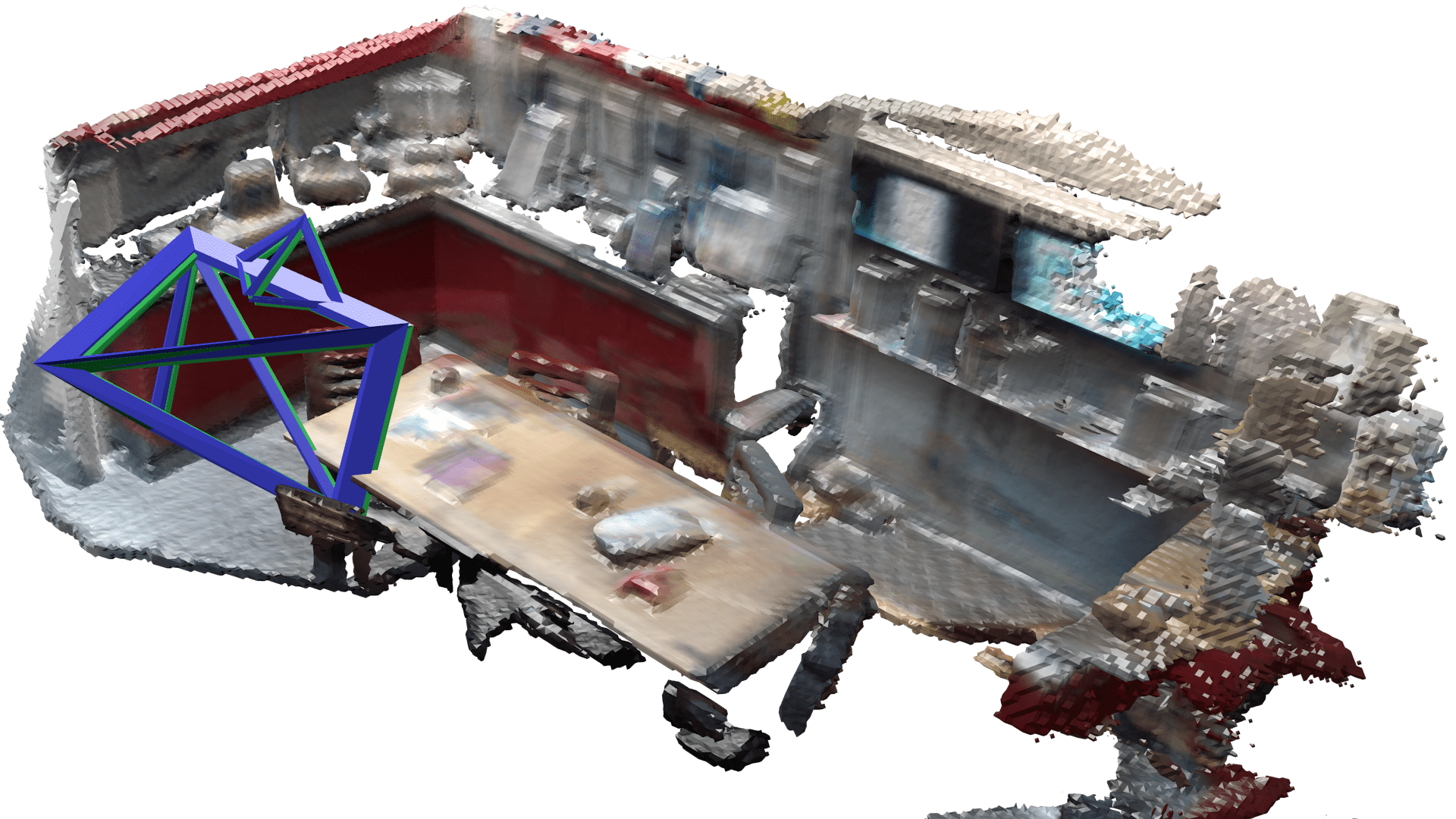}
     \put(0, 5){\#2}
    \end{overpic}
    \begin{overpic}[width=0.32\textwidth]{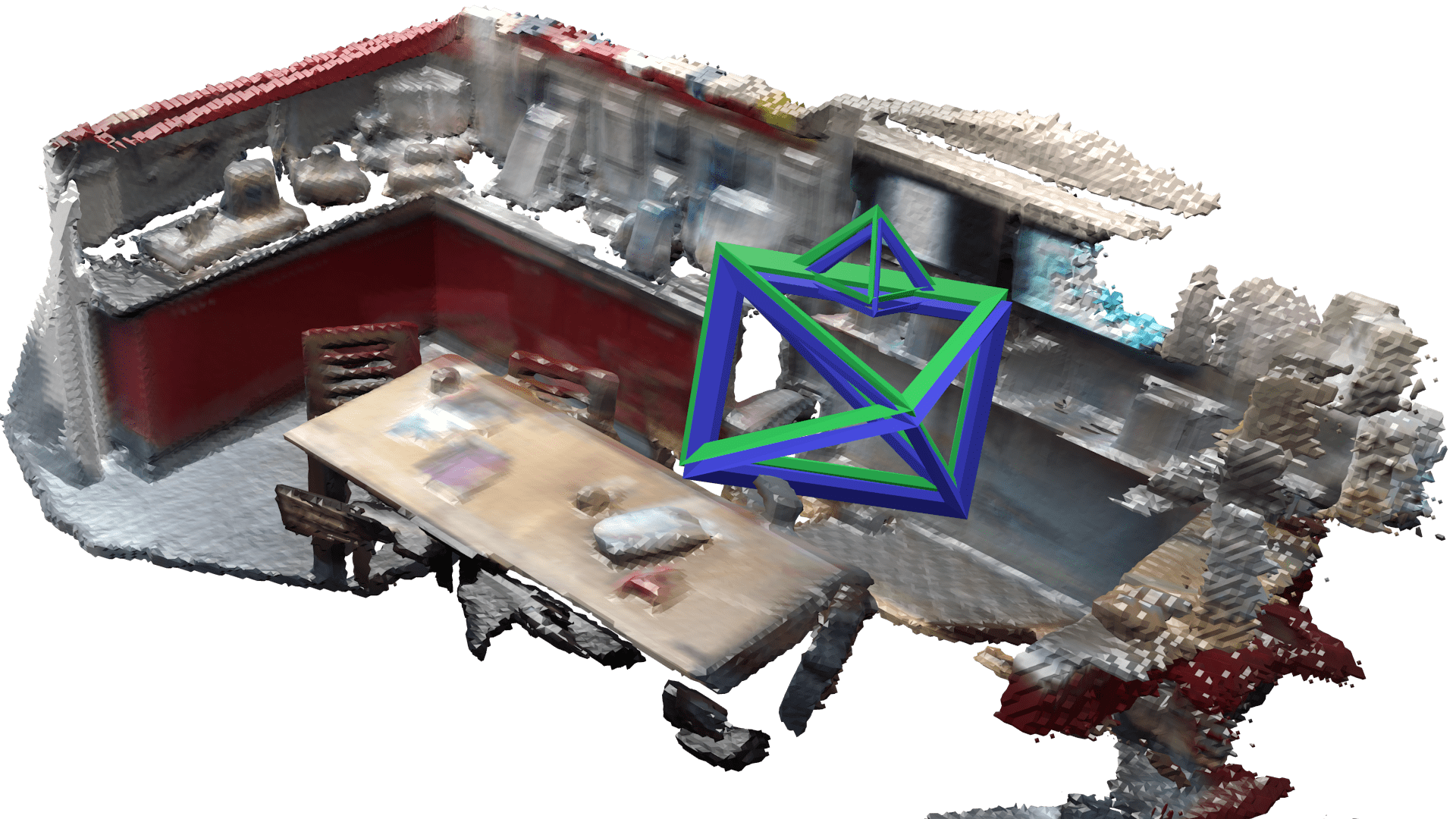}
      \put(0,5){\#3}
    \end{overpic}
    \\
    \begin{overpic}[width=0.32\textwidth]{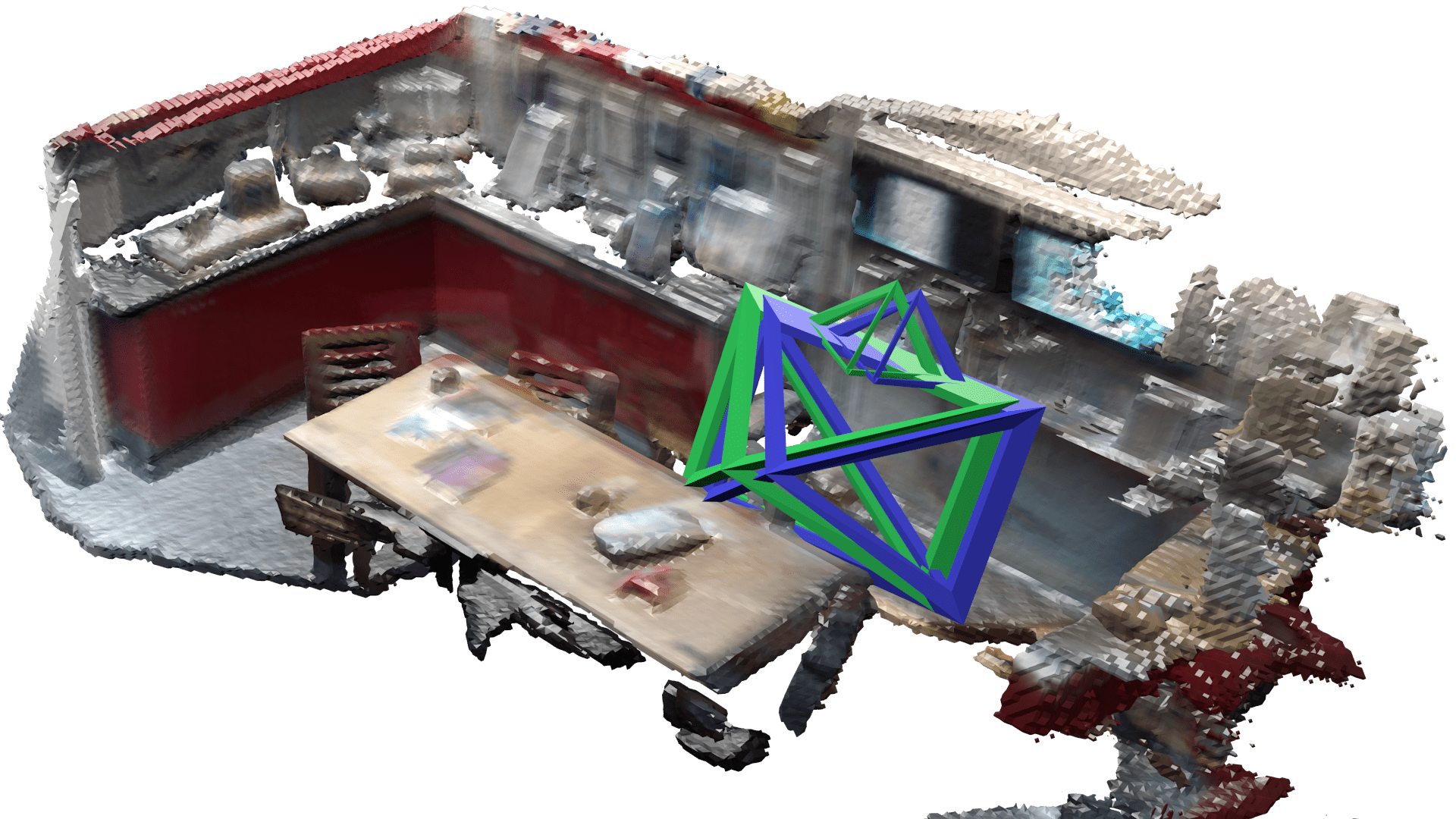}
      \put(0,54){Median}
      \put(0,5){\#2499}
    \end{overpic}
    \begin{overpic}[width=0.32\textwidth]{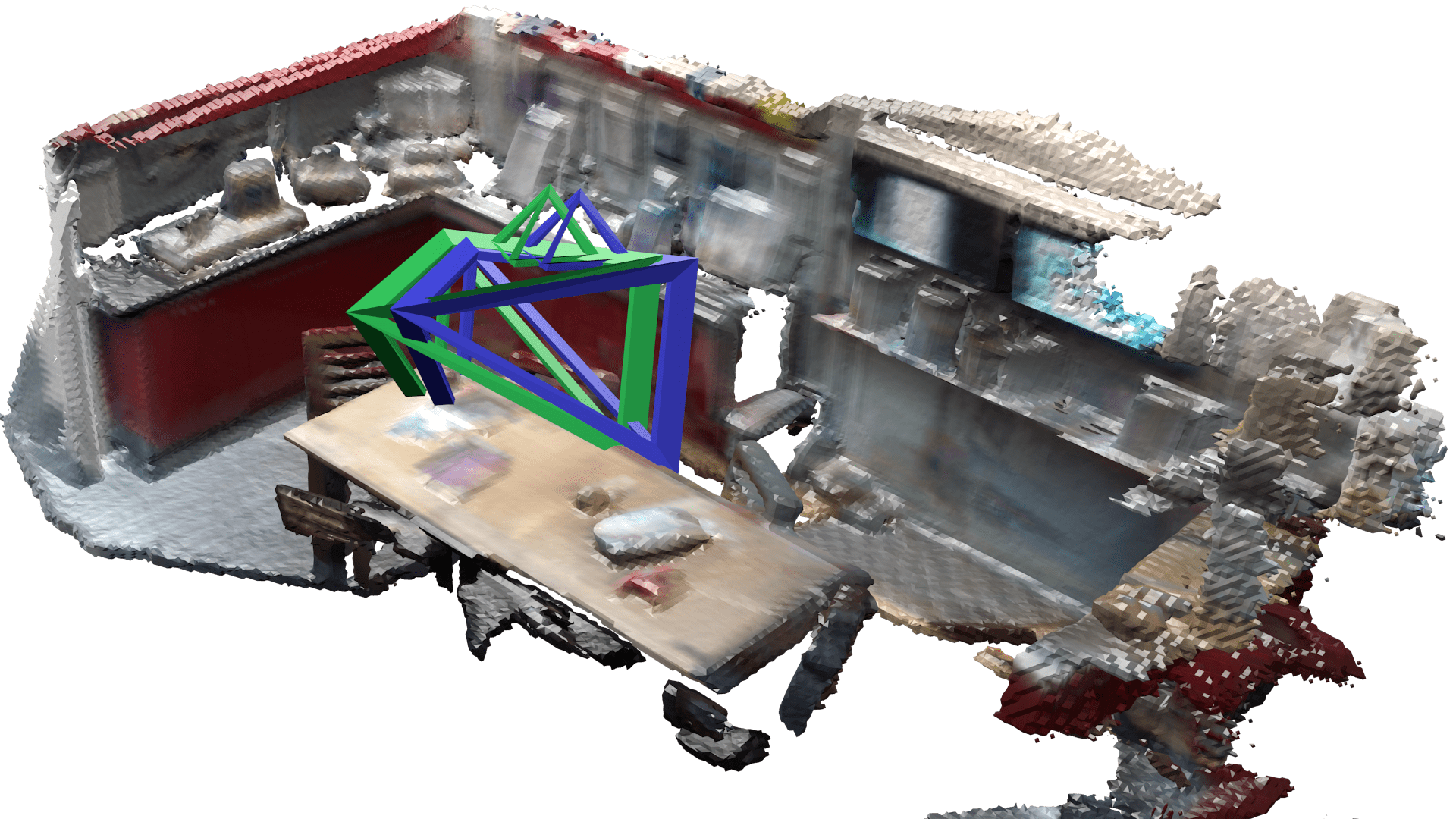}
      \put(0,5){\#2500}
    \end{overpic}
    \begin{overpic}[width=0.32\textwidth]{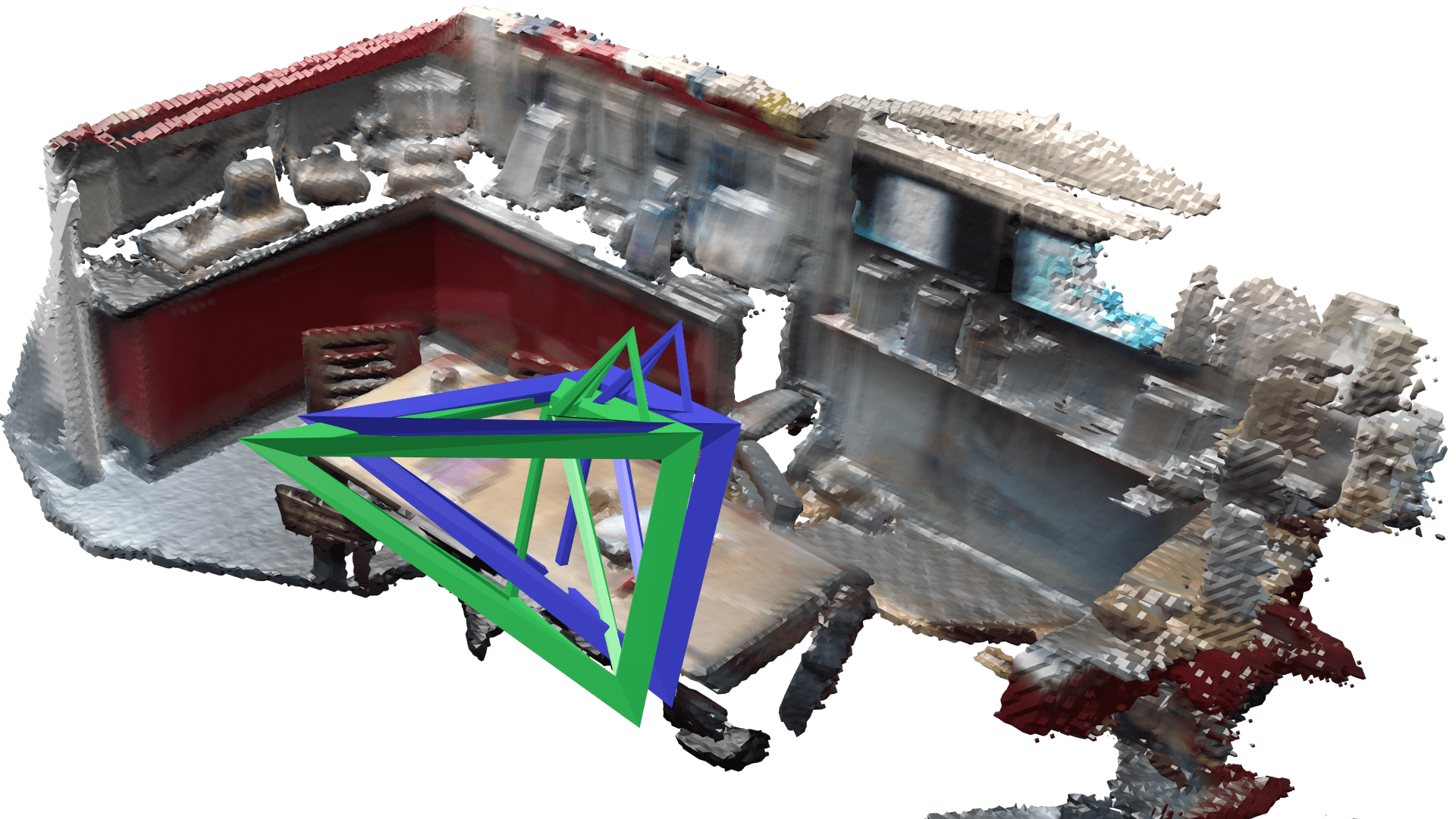}
      \put(0,5){\#2501}
    \end{overpic}
    \\
    \begin{overpic}[width=0.32\textwidth]{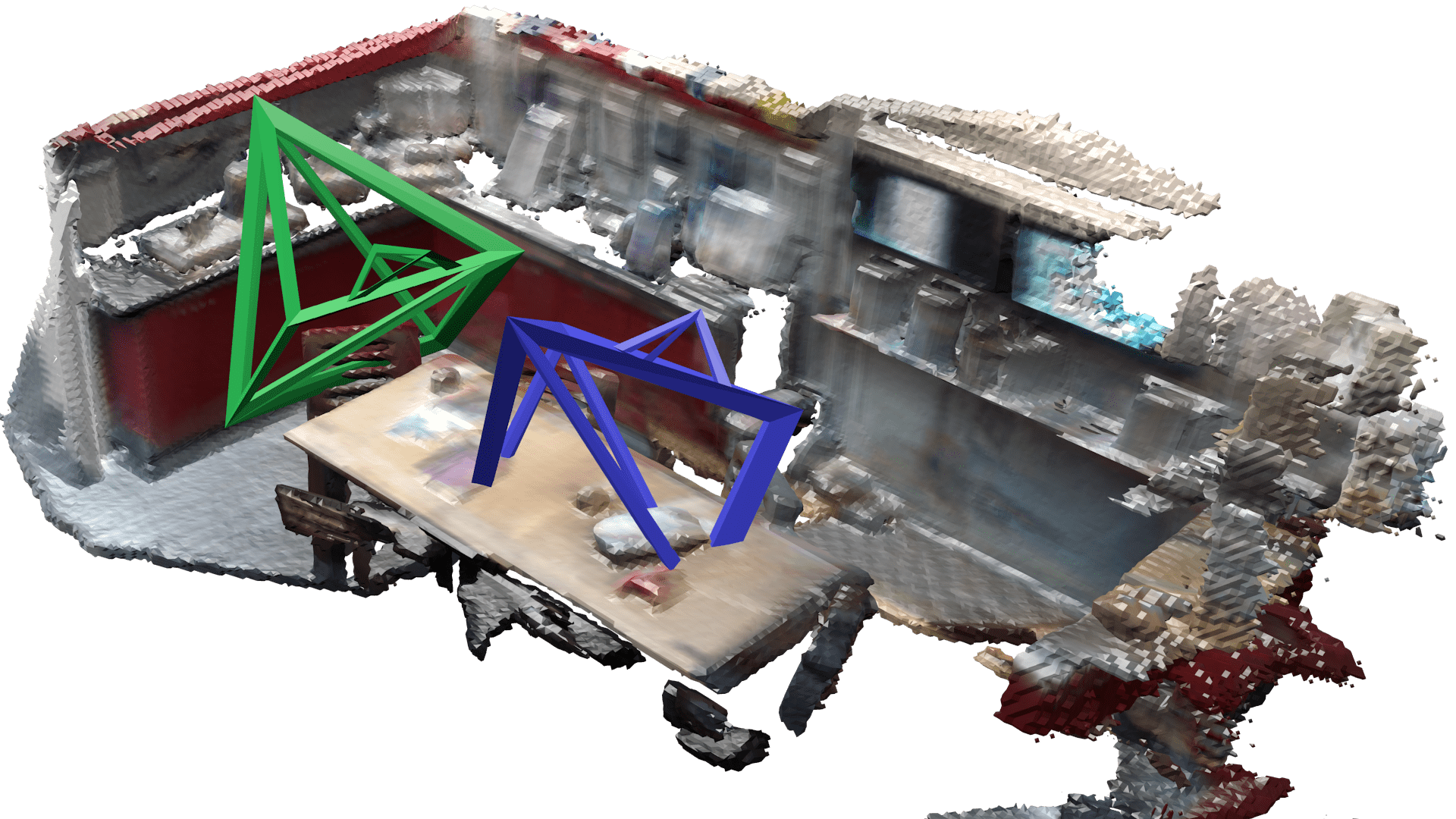}
      \put(0, 54){Worst}
      \put(0, 5){\#4998}
    \end{overpic}
    \begin{overpic}[width=0.32\textwidth]{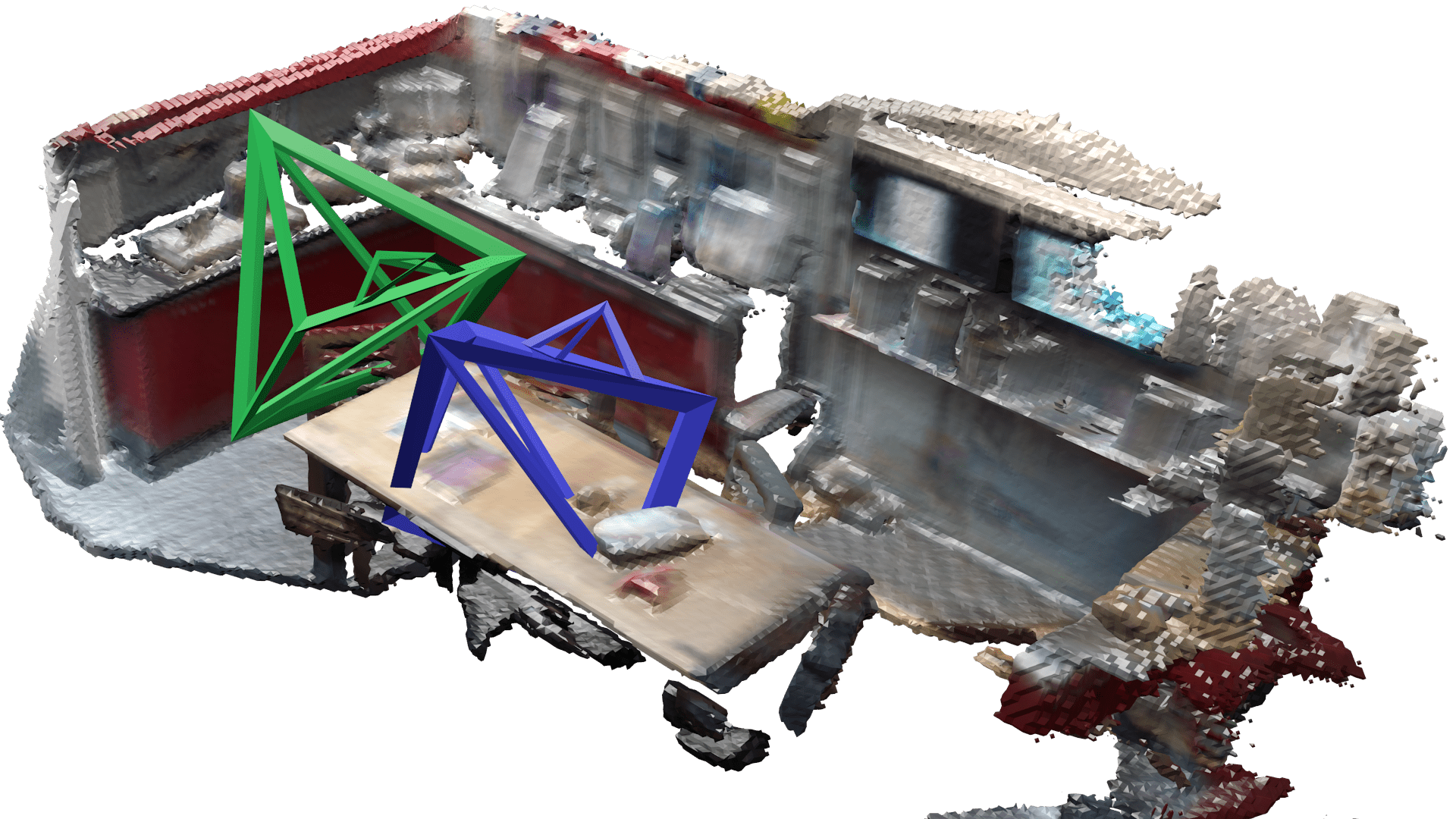}
      \put(0,5){\#4999}
    \end{overpic}
    \begin{overpic}[width=0.32\textwidth]{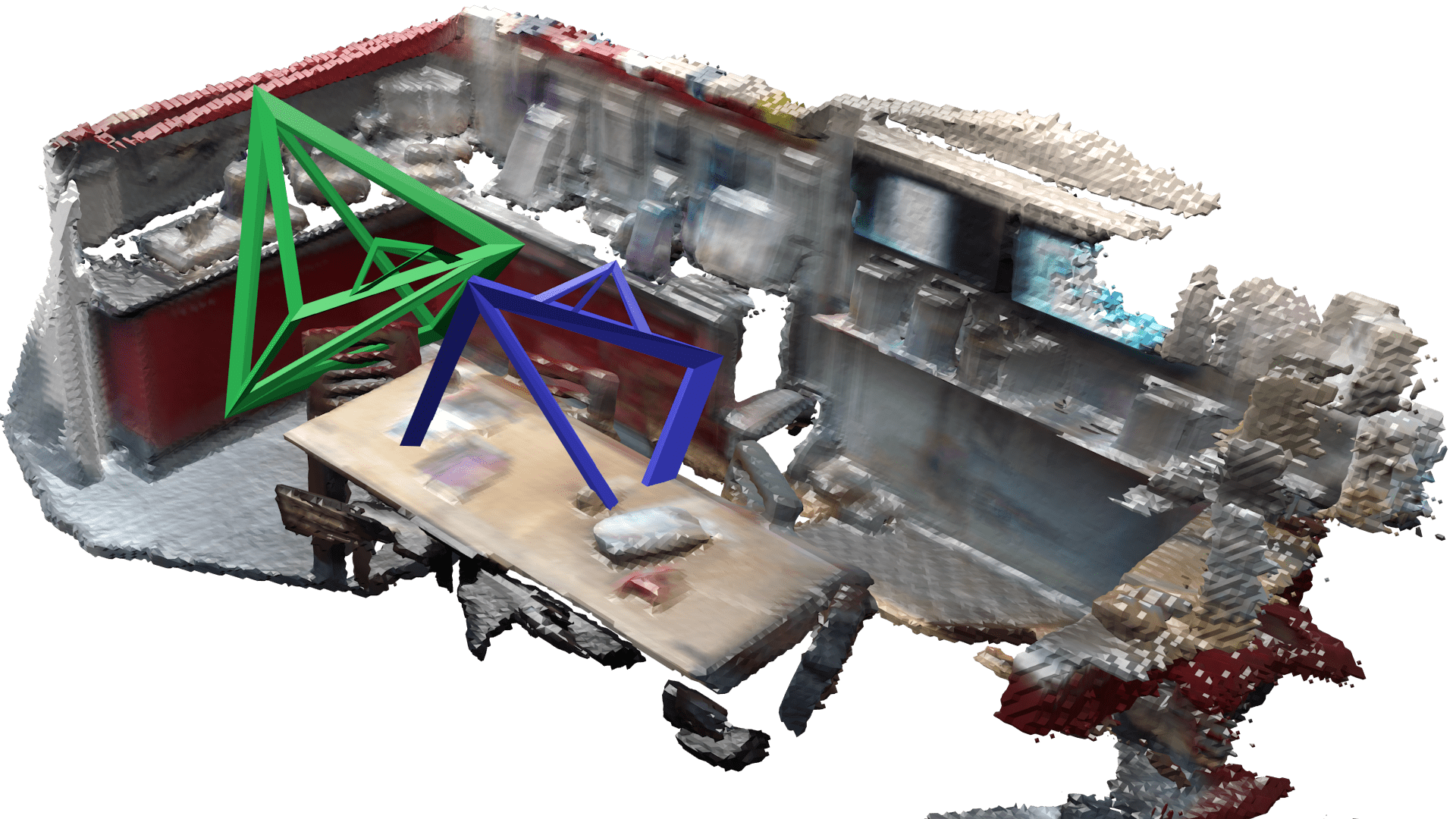}
      \put(0,5){\#5000}
    \end{overpic}
    \caption{\textbf{Qualitative results: Kitchen.} Best, median and worst pose estimates on the \emph{Kitchen} test set, grouped by row. The green camera shows each ground truth pose, while blue cameras show our estimates.}
    \label{fig:qualitative}
\end{figure*}

\section{Strided Sampling}
\begin{table*}[!t]
\centering
\caption{{\bf Strided nearest neighbor selection: 7-Scenes.} When building the fully connected graph of nodes with features representing images, we balance relevance with diversity when selecting nearest neighbors to the query image in the image retrieval embedding space. In this table we show that regularly sampling in the sorted list of nearest neighbors in image retrieval embedding space is better done with a stride, rather than simply taking the $N\!-\!1\!=\!7$ nearest neighbors. Note that on scenes with fewer training sequences and smaller scene sizes, the advantages of diversity might appear negligible, but for larger scenes, such as Kitchen, the improvements in performance are rather significant.}\label{t:strided}
\small
\setlength\tabcolsep{2pt} 
\begin{tabular}{lccccccccc|cc}
\midrule
 & stride ($K$) & Chess & Fire & Heads & Office & Pumpkin & Kitchen & Stairs &&& Avg. \\
\toprule
\mbox{Ours w/o stride}
  & 1
  & \tableentry{\textbf{0.09}}{2.88} 
  & \tableentry{0.26}{9.93} 
  & \tableentry{0.16}{13.76} 
  & \tableentry{\textbf{0.16}}{4.94} 
  & \tableentry{0.17}{3.57} 
  & \tableentry{0.25}{5.19} 
  & \tableentry{0.27}{10.19} 
  &&& \tableentry{0.19}{7.21} 
  \\
  \midrule
    Ours
  & 5
  & \tableentry{\textbf{0.09}}{\textbf{2.85}} 
  & \tableentry{\textbf{0.24}}{\textbf{9.19}} 
  & \tableentry{\textbf{0.15}}{\textbf{12.63}} 
  & \tableentry{\textbf{0.16}}{\textbf{4.59}} 
  & \tableentry{\textbf{0.16}}{\textbf{3.58}} 
  & \tableentry{\textbf{0.21}}{\textbf{4.33}} 
  & \tableentry{\textbf{0.25}}{\textbf{8.48}} 
  &&& \tableentry{\textbf{0.18}}{\textbf{6.52}} 
  \\

\bottomrule
\end{tabular}
\end{table*}

To construct a fully connected graph, we have the following strategy. For each anchor training image in the training stage, or each query image in the test stage, we retrieve the $(N\!-\!1)K$ most similar images in the embedding feature space. 
We then randomly choose an offset $0\!\leq\!k\!<\!K$ and sub-sample the nearest neighbors with regular intervals yielding every $K$-th image in the ordered list of similar images. In our experiments, we set $N\!=\!8$, and $K\!=\!5$ for 7-Scenes and $K\!=\!3$ for Cambridge Landmarks. This method allows us to favorably balance similarity and diversity of images from the training set. See Figure~\ref{fig:strided}. We illustrate the effectiveness of our strided nearest neighbor selection strategy in Table~\ref{t:strided}. We compare our method against a model lacking the strided nearest neighbor selection. Results clearly reflect that it is useful, especially for scenes with large view coverage \eg~\textit{Kitchen} or \textit{Stairs}.

\begin{figure*}[!h]
    \centering
    \begin{overpic}[width=0.12\textwidth]{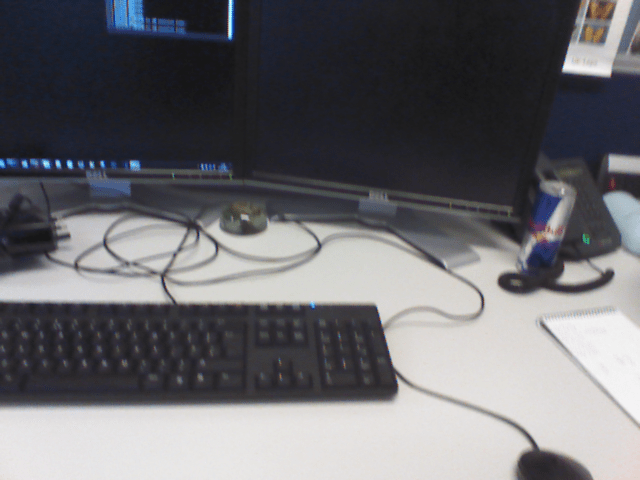}
    \end{overpic}
    \begin{overpic}[width=0.12\textwidth]{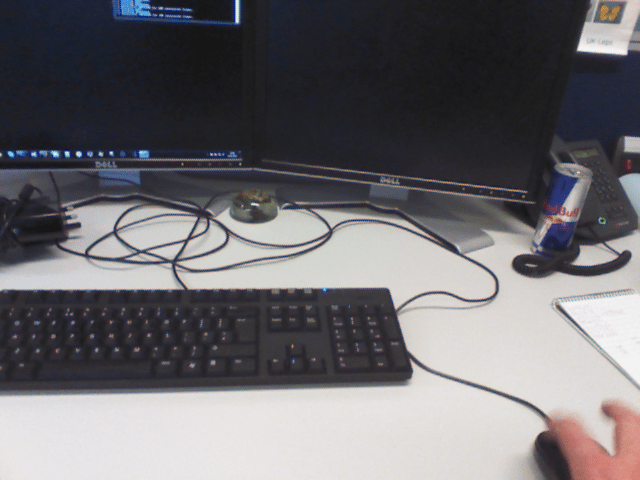}
    \end{overpic}
    \begin{overpic}[width=0.12\textwidth]{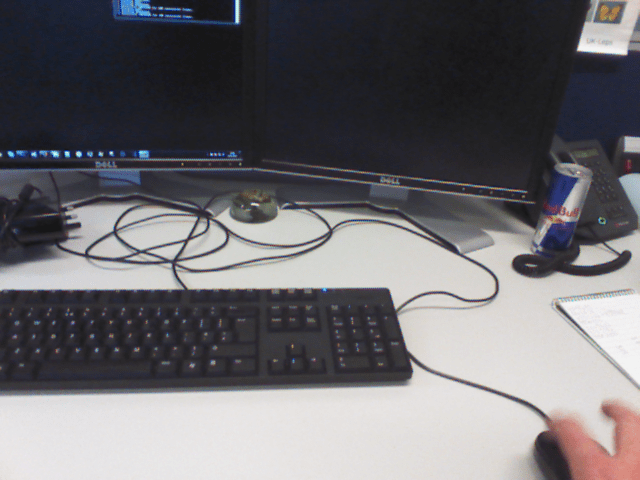}
    \end{overpic}
        \begin{overpic}[width=0.12\textwidth]{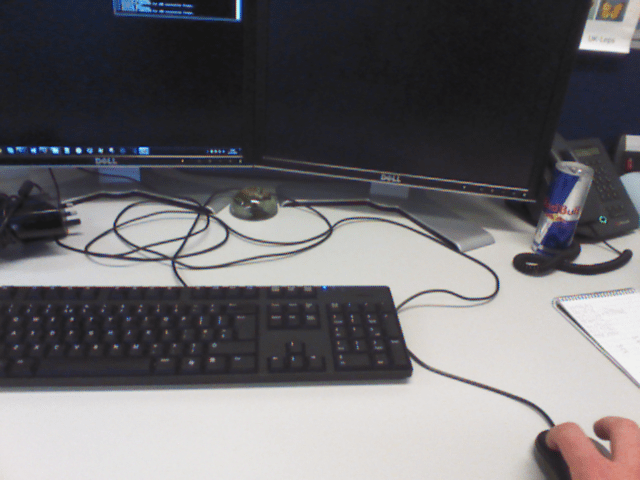}
    \end{overpic}
        \begin{overpic}[width=0.12\textwidth]{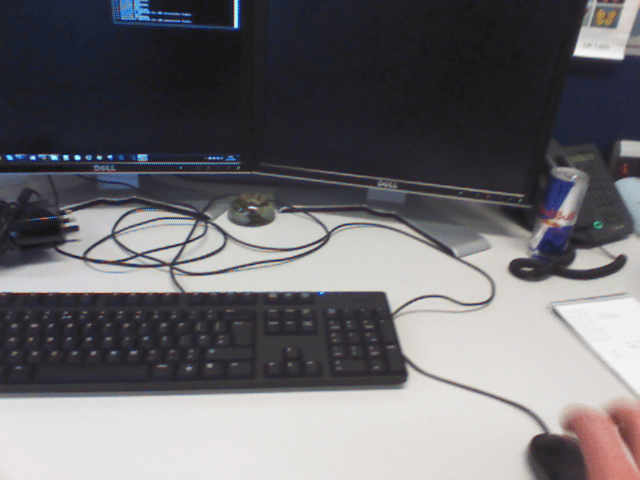}
    \end{overpic}
        \begin{overpic}[width=0.12\textwidth]{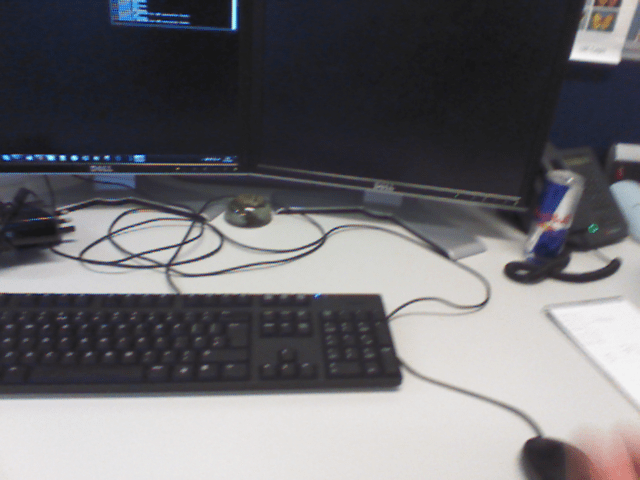}
    \end{overpic}
        \begin{overpic}[width=0.12\textwidth]{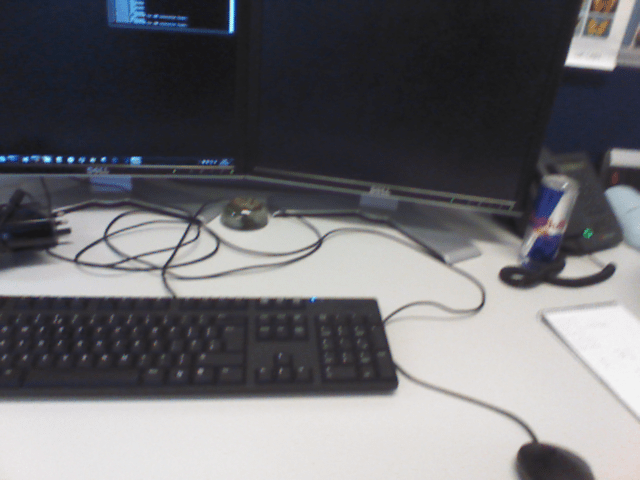}
    \end{overpic}
        \begin{overpic}[width=0.12\textwidth]{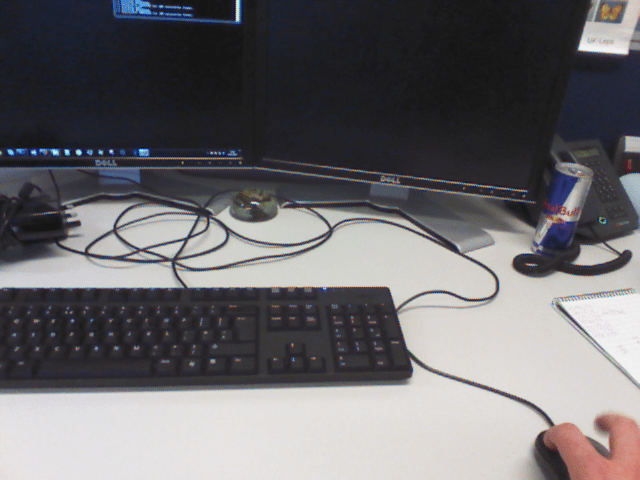}
    \end{overpic}
    \\
    \begin{overpic}[width=0.12\textwidth]{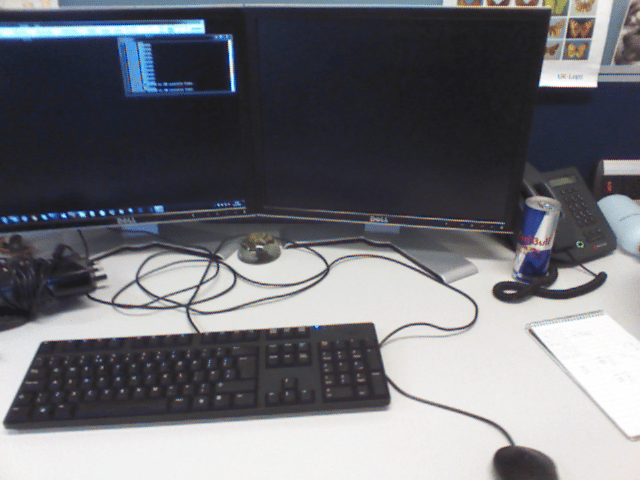}
    \end{overpic}
    \begin{overpic}[width=0.12\textwidth]{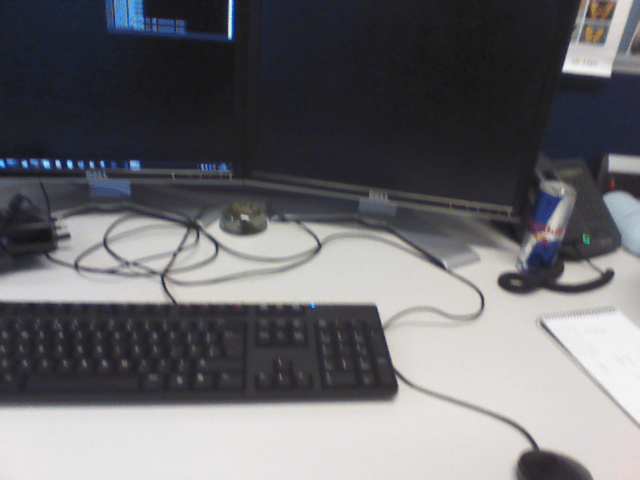}
    \end{overpic}
    \begin{overpic}[width=0.12\textwidth]{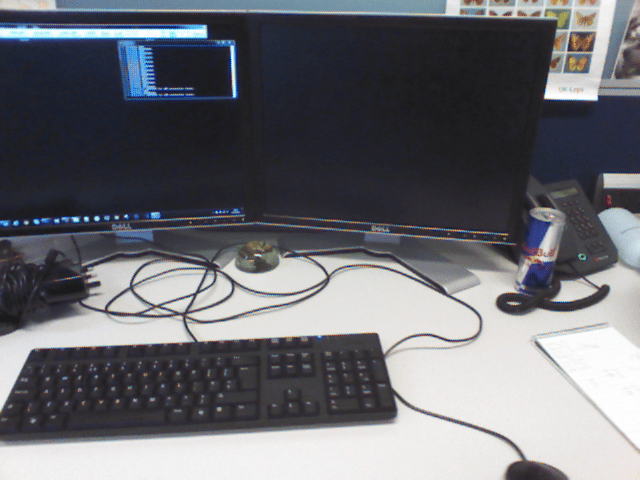}
    \end{overpic}
        \begin{overpic}[width=0.12\textwidth]{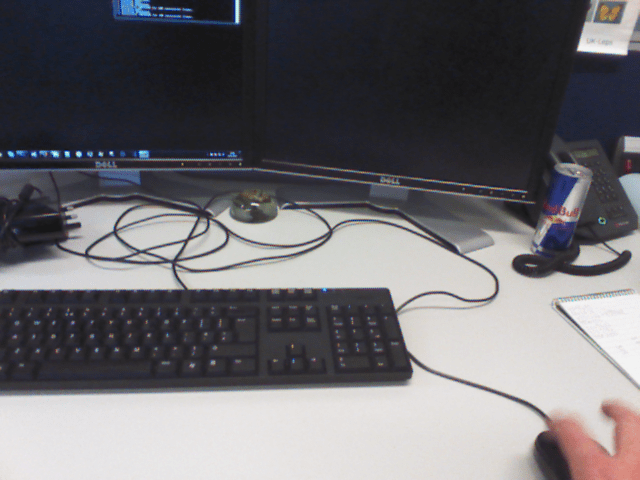}
    \end{overpic}
        \begin{overpic}[width=0.12\textwidth]{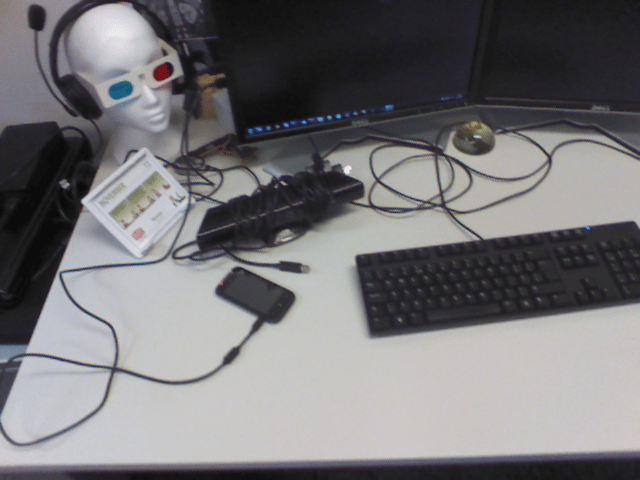}
    \end{overpic}
        \begin{overpic}[width=0.12\textwidth]{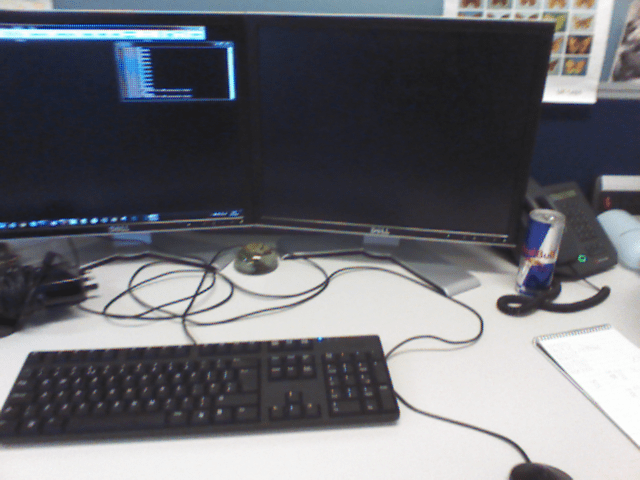}
    \end{overpic}
        \begin{overpic}[width=0.12\textwidth]{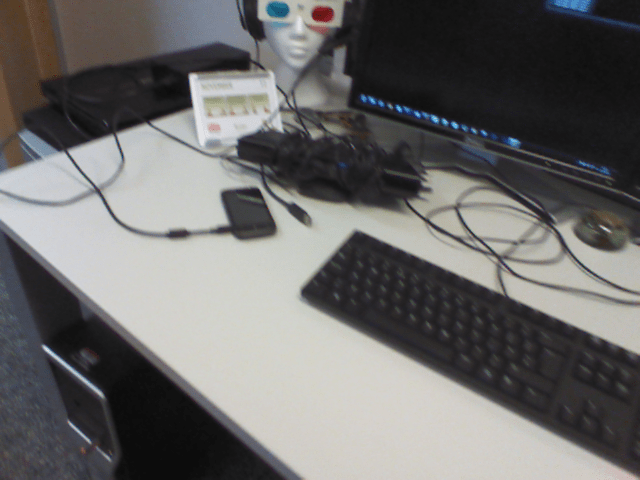}
    \end{overpic}
        \begin{overpic}[width=0.12\textwidth]{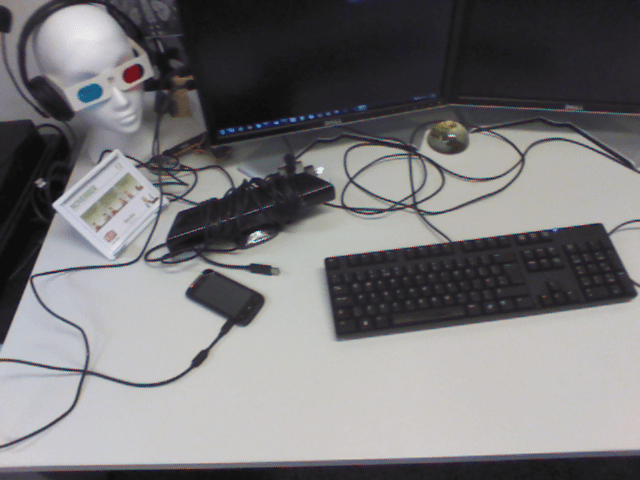}
    \end{overpic}
  
    \caption{\textbf{Strided nearest neighbor selection: Heads.} Example $N=8$ images for building a fully connected graph during training. The first column shows anchor training images. The $1^{st}$ row shows images sampled without striding, $K=1$. The $2^{nd}$ row shows images sampled with stride $K=5$. Our sampling strategy generates much more diversity in viewpoints compared to baseline ($1^{st}$ row) while keeping the relevance of content.}
    \label{fig:strided}
\end{figure*}

\section{Number of Recursions}

We investigate the effect of changing the number of iterations of message passing in our graph neural network (GNN).
We train our model with different numbers of iteration values $R\!\in\!{1,2,3,4}$ for $100$ epochs on the \textit{Pumpkin} scene of the 7-Scenes dataset, and compare the models' test performance on scenarios corresponding to the diagonal entries in the following inset table:

{
\renewcommand{\arraystretch}{1.2}
\vspace{1em}
\begin{tabular}{c|cccc}
     \toprule
     \backslashbox{Train}{Test} & $R=1$ & 2 & 3 & 4 \\
     \hline
     $R=1$ &  -2.24   &  &  &\\
     $R=2$ &  -0.71 & -2.34 & -0.85 & -0.88 \\ 
     $R=3$ &  & &  \textbf{-2.36} & \\
     $R=4$ &  & &  & -2.31 \\
     \bottomrule
\end{tabular}
     \label{t:numiter}

For the performance measure, lower is better. We use the optimization objective as given by
\begin{equation}
    L = \frac{1}{N_e} \sum_{v_{ij} \in E} d(\hat{p}_{ij},p_{ij}) \ .
\end{equation}
Here, $N_e$ is the number of edges, $d(\cdot)$ is the distance function between the predicted relative camera pose $\hat{p}_{ij}$ and the ground truth relative camera pose $p_{ij}$. The distance function $d$ is defined as 
\begin{align}\label{eq:loss}
    d(\hat{p}_{ij},p_{ij}) &= |\hat{t}_{ij}- t_{ij}| e^{-\beta_0} \nonumber\\
    &+ \beta_0 + |log(\hat{q}_{ij})- log(q_{ij})| e^{-\gamma_0}\!+ \gamma_0 \ ,
\end{align}
where $\beta_0=0$ and $\gamma_0=-3$.

The results show that a GNN with $R\!=\!3$ recursion works best, followed by $R\!=\!2$. One can also see, that a single pass of message passing is not enough for node representations to collect sufficient information from their surroundings.
We further our experiment with fixing the number of recursions $R\!=\!2$ at training time and varying it $R\!\in\!{1,2,3,4}$ during test time. 
The results are presented on the second row of the inset table above.
We can conclude that the performance of our GNN significantly degrades if $R$ differs between the train and test stages.

\section{Saliency Visualization}
\label{sec:saliency}

To further investigate representations learned by our approach and visualize the difference to the baselines, we generate saliency maps for exemplary test images. See \Cref{fig:saliency}. We compare our method against \emph{Baseline-1} (w/o GNN) and \mbox{AtLoc~\cite{wang2020atloc}}. Saliency maps are a convenient way to gain some insight into the operation of a CNN, by making explicit which image pixels contribute strongly to the model's predictions.
Saliency is computed by taking the gradient of the prediction \wrt the input. In the first row of \Cref{fig:saliency}, one can see that \emph{Baseline-1} focuses on a featureless region (gray table), which, despite being a rigid scene object, is hardly suitable for pose estimation. AtLoc's focus is rather scattered, including some geometrically meaningful regions like the tips on the headphones, but also many featureless regions, \eg the white wall on the left. On the contrary, our proposed method ignores most featureless regions and directs its attention to geometrically meaningful parts such as the corners of the phone box. Its saliency map is also more concentrated on sharp local features. 
In the second row, \emph{Baseline-1} and AtLoc both heavily rely on lighting artifacts on the stairs, which obviously is not a robust geometric feature. In contrast, the proposed GNN method largely ignores the lighting pattern and focuses on the edges of the handrail. We find this an interesting example that suggests our method generalizes well and does not heavily focus on scene-specific features.

\begin{figure}[!h]
    \centering
    \renewcommand\tabcolsep{0pt}
    \small
    \begin{tabular}{ccc}
\includegraphics[width=0.33\columnwidth]{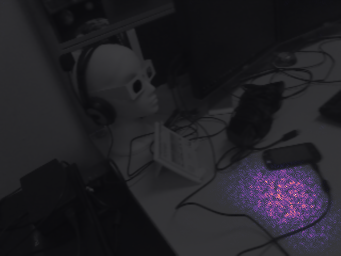} &
\includegraphics[width=0.33\columnwidth]{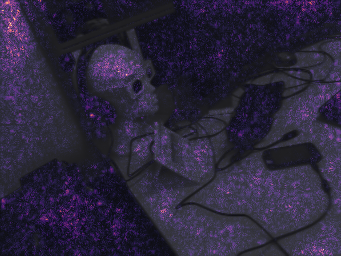} &
\includegraphics[width=0.33\columnwidth]{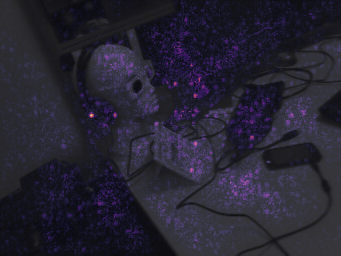} \\
\includegraphics[width=0.33\columnwidth]{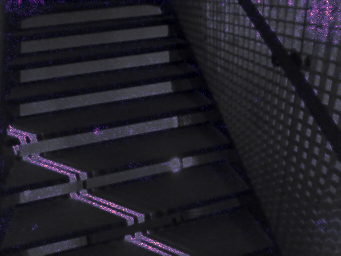} &
\includegraphics[width=0.33\columnwidth]{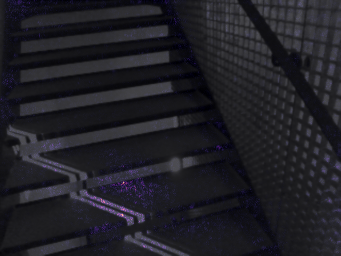} &
\includegraphics[width=0.33\columnwidth]{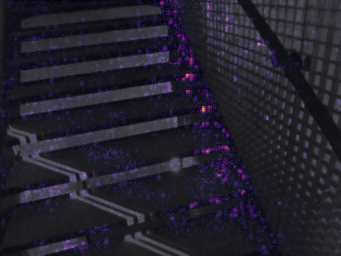} \\
 Baseline-1 & AtLoc~\cite{wang2020atloc} & Ours \\
\end{tabular}
\caption{Saliency maps for test images in \emph{Heads} ($1^{st}$ row) and \emph{Stairs} ($2^{nd}$ row) scenes. A saliency map offers a visualization of pixels that contribute the most to predictions. See Section~\ref{sec:saliency}.}
\label{fig:saliency}
\end{figure}

\section{Reproducibility Challenges}

\subsection{GL-Net \cite{glnet}}
The authors of \mbox{GL-Net \cite{glnet}} kindly sent us the two implementation files of their network architecture. However, despite our best efforts we have not been able to reproduce their results on 7Scenes. The training scheme the paper describes together with the network architecture in the files they sent always resulted in the training procedure over-fitting to the training data (high validation loss).
Our repeated e-mails asking for the full training code remained unanswered. \\
The code link on the first author's research page\footnote{\url{https://feixue94.github.io}} is blank, the project page link\footnote{\url{https://github.com/feixue94/grnet}} in the paper's abstract is \emph{empty} since the paper's publication date 6/13/2020.
In contrast to earlier works from the first author of \mbox{GL-Net~\cite{glnet}}, we are hopeful that the code or enlightening instructions will eventually be published to allow more accurate comparisons of their otherwise inspiring method.
We've compared to their image retrieval strategy (the next images in the sequence) using our network architecture in \mbox{Table~4 row 3} in the paper. It was not surprising to see, that methods using image retrieval outperform their sequential strategy.
\ifmyarxiv
\else
Please see our attempts in \emph{ICCV21\_7821\_code/glnet} in the supplementary zip.
\fi

\subsection{AnchorPoint}

The official code published for \mbox{AnchorPoint \cite{saha2018improved}} seems to only regress 4DoF camera pose and only compute the test translation error in 2D\footnote{\url{https://github.com/Soham0/Improved-Visual-Relocalization/blob/8eacdc9d09054ec2f086c0668c83dc0af324ccae/localize_scene.py\#L367}}. 
Despite unaddressed reproducibility issues\footnote{\url{https://github.com/Soham0/Improved-Visual-Relocalization/issues/1}}, we've been able to run the code on the Cambridge dataset, as can be seen in \Cref{t:cambridge} ``AnchorPoint reprod.''. 
In the paper and in \Cref{t:cambridge} we reported numbers with \emph{\_2D} to denote that they only measure a 2D Euclidean distance between the predicted camera pose and the ground truth camera pose.
\ifmyarxiv
\else
Please see our attempts at \emph{ICCV21\_7821\_code/anchorPoint} in the supplementary zip.
\fi

\subsection{CamNet}

We have not been able to reproduce results with the official code for \mbox{CamNet \cite{camnet}}. 
Environment files and custom network modules are missing from the uploaded source. For instance, they use the \emph{torchE} package; however, we could not find that package online. 
After fixing bugs we managed to train the model but we observed the models to be over-fitting to the training data.
We share our concerns with others in the community\footnote{\url{https://github.com/dingmyu/CamNet/issues/7}}. We sent the authors queries for advice on how to proceed, but they did not reply.
\ifmyarxiv
\else
Please see our attempts in \emph{ICCV21\_7821\_code/camnet} in the supplementary zip.
\fi

\else
\fi

\end{document}